\documentclass{article} % For LaTeX2e
\usepackage{iclr2020_conference,times}
\usepackage{authblk}
\usepackage{amsmath,amsfonts,bm}

\def\eqref#1{equation~\ref{#1}}
\def\1{\bm{1}}

\DeclareMathAlphabet{\mathsfit}{\encodingdefault}{\sfdefault}{m}{sl}
\SetMathAlphabet{\mathsfit}{bold}{\encodingdefault}{\sfdefault}{bx}{n}

\let\ab\allowbreak

\usepackage{hyperref}
\usepackage{url}
\usepackage[utf8]{inputenc} % allow utf-8 input
\usepackage[T1]{fontenc}    % use 8-bit T1 fonts
\usepackage{hyperref}       % hyperlinks
\usepackage{url}            % simple URL typesetting
\usepackage{booktabs}       % professional-quality tables
\usepackage{amsfonts}       % blackboard math symbols
\usepackage{nicefrac}       % compact symbols for 1/2, etc.
\usepackage{microtype}      % microtypography

\usepackage{rotating,multirow}
\usepackage{graphicx,placeins}
\usepackage{amsmath,amsthm,booktabs}
\usepackage{thm-restate}
\usepackage{soul}
\usepackage{capt-of}

\usepackage{bm}
\usepackage{array}
\newcolumntype{x}[1]{>{\centering\arraybackslash\hspace{0pt}}p{#1}}

\usepackage{diagbox}

\usepackage{tikz}
\usepackage{tikz-cd}
\usetikzlibrary{matrix,calc,arrows}
\usetikzlibrary{tikzmark}

\tikzstyle{b} = [rectangle, draw, fill=blue!20, node distance=3cm, text width=6em, text centered, rounded corners, minimum height=4em, thick]
\tikzstyle{c} = [rectangle, draw, inner sep=0.5cm, dashed]
\tikzstyle{l} = [draw, -latex',thick]

\title{Mask Based Unsupervised Content Transfer}

\iclrfinalcopy

\author[1]{Ron Mokady}
\author[1]{Sagie Benaim}
\author[1,2]{Lior Wolf}
\author[1]{Amit Bermano}
\affil[1]{The School of Computer Science, Tel Aviv University}
\affil[2]{Facebook AI Research}
\begin{document}

\maketitle

\begin{abstract}
    We consider the problem of translating, in an unsupervised manner, between two domains where one contains some additional information compared to the other. The proposed method  disentangles the common and separate parts of these domains and, through the generation of a mask, focuses the attention of the underlying network to the desired augmentation alone, without wastefully reconstructing the entire target.
    This enables state-of-the-art quality and variety of content translation, as demonstrated through extensive quantitative and qualitative evaluation. Our method is also capable of adding the separate content of different guide images and domains as well as remove existing separate content. Furthermore, our method enables weakly-supervised semantic segmentation of the separate part of each domain, where only class labels are provided. Our code is available at \url{https://github.com/rmokady/mbu-content-tansfer}.

\end{abstract}

\section{Introduction}

\begin{figure}
\centering
%\begin{tabular}{c}
  %\includegraphics[width=0.5\linewidth, clip]{results/glasses_2.jpg} &
  %\hspace{2.2cm} \includegraphics[width=0.795\linewidth, clip]{results/glasses_orientation_1.jpg} \\
  %\includegraphics[width=0.95\linewidth, clip]{results/glasses_orientation_2.jpg} \\
  %\end{tabular}
%   \includegraphics[width=0.99\linewidth, clip]{results/glasses_orientation_v2.jpg} 
    \includegraphics[width=0.99\linewidth, clip]{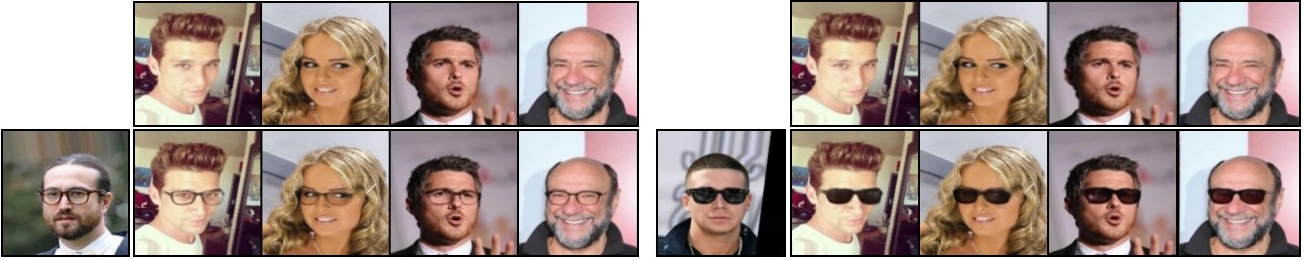} 
\caption{Content transfer example. Given an image of a face with glasses (left), and another image of a face without ones (top), the proposed method successfully identifies and translates the specified glasses from the former domain to the latter one. }
  \label{fig:orientation}
\end{figure}

The task of content transfer, as depicted in Fig.~\ref{fig:orientation}, involves identifying the component of interest (for example, glasses) in a given input (for example, an image of a face), adapting it, and adding it to a second given input (for example, another image of a face, without glasses), hopefully in the semantically correct manner. Such an operation can be used to prototype or demonstrate changes in appearance \cite{gatys2016image}, augment music \cite{Grinstein18Audio}, compose text \cite{prabhumoye2019towards}, generate data for training purposes \cite{mueller2018ganerated}, etc.

Recent advancements (\cite{munit, DRIT_plus}) translate one domain to another with varying styles, but not content. Others (\cite{lample2017fader, attgan, stgan}) produce images in a target domain with a given attribute (e.g glasses), but such attribute is unique and not varying or controlled (e.g., not allowing the specification of a specific pair of glasses).

A recent advancement in the realm of attribute transfer has been presented by \cite{ori}. In this work, the input is two domains of images, such that the images in one domain, $B$, contain a specific class (e.g. faces with facial hair), while in the other domain, $A$, the images do not (e.g., faces without facial hair). Training on this input, the method learns to transfer only the specific class information from an unseen image in the domain $B$ to an unseen one in the domain $A$, while preserving all other details. The proposed architecture yields a simple network which is able to perform the required disentanglement through an emergence effect. In this setting, the content to be added by the system is not explicitly marked in the target domain, nor does it have an equivalent counterpart in the source domain for training (e.g. an image with and without glasses of the same person). A form of weak supervision can be provided by a simple annotation of whether the relevant content exists or not in every example. However, this method, and other typical ones addressing similar tasks, generate the details for the entire image, by using auto-encoder or GAN-based architectures, resulting in a degradation of details and quality.

%implicitly, such that it emerges due to the laziness principle \lw{NOT SURE WHAT THIS IS}\ab{I'm trying to explain Ori's Lemma 2 in one word. Is it not all made possible basically because the network learning process is lazy, striving for minimal change that will reduce the loss? Please feel free to change to something more correct}\lw{CALL IT EMERGENCE}\sab{The laziness occurs due the third "zero" loss, essentially with the zero loss the decoder MUST learn face features from the common encoding. Assuming this, the "easiest" thing for the decoder to do in order to decode the image with glasses is to add only the glasses from the separate encoding}. 

In this paper, we build upon the emerging disentanglement idea, but also adopt the growing understanding that one should minimize redundant use of computational resources and model parameters (\cite{chen2016attention,atenguided,atengan}), to the aforementioned task. 
%\lw{I DO NOT SEE HOW WE MINIMIZE USAGE OF SUCH RESOURCES} 
In other words, using a mask, we focus the attention of the baseline network to the desired augmentation alone, without asking it to wastefully reconstruct the entire target. As can be seen in Fig~\ref{fig:overview}, the method consists of two main steps. The first is the disentanglement step, which encodes the domain specific and the domain invariant contents separately, and is inspired by the work of \cite{ori}. The second step is the key insight of our proposal. It locates the part of the target that should be changed and generates relevant augmentation content to go with it. This allows keeping the unrelated details intact, facilitating a great improvement in generation quality; The augmentation focuses on the relevant part, leaving all other details to be taken from the target image directly, without going through the bottleneck of an auto-encoder-like module.

By applying this simple yet effective principle and a novel regularization scheme, our method preserves target details which are irrelevant to the augmentation and is able to improve upon the state-of-the-art in terms of quality and variety of the content transferred. Furthermore, we demonstrate how the method performs well, even {when presented with images outside the domain trained on}, and that the aforementioned mask, generated by the system to mark the regions of augmentation, is accurate enough to provide a semantic segmentation of the content transferred. 

Lastly, our method can also be used in the opposite direction --- to remove the domain specific attribute, and thus translating from $B$ to $A$. Despite not being our main focus, we outperform the various literature methods that address this task (\cite{lample2017fader,stgan,attgan,ori}). Our advantage is also evident in that once an object is removed, we can then add a different attribute to the resulting image, thus allowing the translation from any two domains, each having a domain specific attribute (e.g removing glasses and adding the smile of a guide image). A related unique ability is that of removing and adding the same attribute, e.g. replacing one's facial hair with a different facial hair.

\section{Previous Work}

%\sab{In the image to image translation problem, the learner is given two domains of visual samples $A$ and $B$, and is asked, given image $a \in A$ to generate an analogue image in domain $B$. In the supervised case, pairs of images are provided during training. \cite{pix2pix} uses a U-net like architecture and and adversarial procedure to produce a suitable mapping. \cite{pix2pixhd} extended this solution to produce high resolution photo-realistic solution. These methods use costly supervision which we do not employ. }

In the unsupervised image to image translation, the learner is given two unpaired domains of visual samples, $A$ and $B$, and is asked, given image $a \in A$ to generate an analogue image in the domain $B$. %\sab{In the unsupervised case, matching pairs are not provided during training.} 
This problem is inherently ill-posed, as multiple analogous solutions may exist. %Different approaches alleviate this ambiguity, by forcing constraints that produce a particular solution: In DTN~\cite{02200}, a pre-trained face embedding is used during training to ensure a consistent mapping. 
In several of different approaches (\cite{CycleGAN2017, discogan, dualgan}) a circularity constraint is used to reduce this ambiguity. %enforce that going from one domain and back we can recover the original sample $a$, %. DistanceGAN~\cite{distgan} enforces the distance between to samples in preserved after translation to the target domain while 
COGAN~\cite{cogan} and UNIT~\cite{unit} enforce a shared latent representation between the two domains.  %\sab{Other methods enforce a particular constrain on on the translated images such as preserving pixel values \cite{cite1}, pixel gradients \cite{cite2} or pairwise distances \cite{distgan}. 
Unlike our method, these methods produce a single solution per input image $a$.  

Moving from one to one mappings, multiple approaches provide many to many mappings. Supervised multimodal approaches, where paired samples are provided, include BicycleGAN~\cite{zhu2017toward}, which injects random noise $z$ in a generator and enforces an encoder to recover $z$ from the target translation, and MAD-GAN~\cite{madgan}. The latter trains multiple generators to produce aligned mappings, which are distant from each other. These methods require paired samples from both domains ---  a costly supervision, which we do not require. 

\textbf{Guided Multimodal Approaches} MUNIT~\cite{munit}, DRIT~\cite{Lee_2018_ECCV}, and DRIT++~\cite{DRIT_plus} are trained on unmatched images. MUNIT trains two encoders; one captures the content of an image, and another its style, inducing disentanglement. During inference, multiple solutions are produced using the style of a guide image in the target domain. In DRIT (and DRIT++), a cycle constraint is employed, in a setting where the generator of each domain consists of two disentangled encoders, one of which encodes the content and the second the style of the image. For all these methods, different {non symmetrical} architectures of encoders are used to capture the style and content. %MUNIT, for example, uses spatial pooling and adaptive instance normalization for the style encoder, while it uses residual connections for the content encoder. In these methods, the latent dimension of the style encoder is much smaller than that of the content encoder. This asymmetry is crucial in these methods, since otherwise the encoder could capture all the information of the input image, and no disentanglement would take place, and the guidance would become mute. 
In MUNIT, for example, residual connections are used for the content encoder, and global pooling and adaptive instance normalization are used for the style encoder. Hence, the style code is significantly smaller in dimensions than the content one. In our method we employ two encoders as well, but the architecture of these encoders is symmetric, allowing both encoders to capture content in both.%domain specific and domain invariant content. 

The most relevant work to ours is that of \cite{ori}, which also uses the setting in which the samples in domain $B$ contain all the information in domain $A$ and some additional information. Two encoders are used --- the first captures the information that is common between the two domains and the second encodes the unique information of domain $B$. The decoder maps the concatenation of the two encodings into an image in the domain $B$, or, in the case that the second encoding is set to zero, to an image in $A$. Content is transferred between images by mixing the encoding of the former type of one image with the encoding of the latter type of a different image.%To map between domains $A$ and $B$ one applies the decoder onto the common encoding of the source image the separate encoding of the target image. 

In many cases, however, only a local area in the image needs to change during translation. Consider the case where $A$ is images of faces and $B$ is faces with facial hair. For $a \in A$, only the location in which the facial hair is placed in $a$ needs to change. In the method of \cite{ori}, the entire image, including other facial features, is generated from scratch, and as a result, many low level details are lost and the quality of generation is reduced. This is not the case for our method, where outside the generated mask, which denotes the location of the facial hair, the content of the generated image is taken from the input image $a$. This is achieved by employing two decoders, one for domain $A$ and one, with two outputs (raw image and mask) in domain $B$, and by a new set of loss terms.

%We note, that for disentanglement of the common and separate information, requires the latent dimension of the common encoder and separate encoder to be low, which causes the reconstruction to be worse and so many low level details to be lost in generation. Unlike their method, we separate the translation to a disentanglement module and a masking module, which allows as bridge the task of unsupervised content transfer to that of weakly supervised semantic segmentation (in which only class information is available) by considering the mask generated as part of the translation. 

\textbf{Mask Based Approaches\quad} The use of masks is prevalent in a variety of visual tasks. For example, for virtual try on, \cite{viton} uses a supervised human parser network to transfer the desired clothing to a target person. Unlike our method, this method cannot perform general content transfer as it crucially relies on a supervised pose estimator. In the context of style transfer, \cite{ma2018exemplar} uses masks to transfer the style of semantically similar regions from the target to the source image. In the context of image to image translation, the one to one case was addressed by \cite{atengan, atenguided}, where in addition to mapping to the target domain, a mask is learned, to cover only the relevant area in the translation. For example, in the case of mapping from horses to zebras, the mask learns to cover the area of the zebra, which allows the background to be taken from the source image, thus allowing for a much better quality of generation. In our method, we extend this masking (or attention) approach to the one to many (guided) case. Note that while~\cite{atengan, atenguided} learn a mask and then employ it directly to the image, this does not allowthe mask to adapt both target and guide image.

\textbf{Weakly supervised semantic segmentation methods} can be stratified based on the type of supervision used. In the first set of methods (\cite{app1, app2, app3}) a bounding box is used. Other approaches (\cite{app4, app5}) use the entire supervision of the fully supervised approach, but are required to find a segmentation in a single shot. Our approach belongs to a set of methods \cite{app6, app7, yuchau, app8} that use only the class label information to find a segmentation. \cite{app7} use the visual cues arising from peaks in class response maps (local maxima) to generate highly informative regions. \cite{zhou2016learning} use Class Activation Maps (CAMs) extracted from a classifier to obtain discriminative localization. \cite{yuchau} use varying dilation rates to transfer surrounding discriminative information to non-discriminative object locations.  \cite{app8} propagate local discriminative parts to nearby regions that belong to the same semantic entity. { Other methods (\cite{regul1, regul2, regul3}) use regularized losses with different levels of supervision.} Unlike these methods, our main focus is on generating the added part in a way that it is adapted to the placement context for the domain specific information  and yet, our segmentation results are competitive with such methods on this task. %\amit{we need something more here. Maybe "and yet we still see comparable results for the domain specific regions"?. Maybe something like this avenue could be explored further in future work? I mean, if we do something completely different (which is what the last sentence says), why are we talking about this at all?}%existing methods, our method uses disentanglement and learns to generate the both parts separately, thus it capable to perform segmentation for both the common and the separate parts.}

% \textbf{Disentanglement methods}
% InfoGAN \cite{infogan} learns a disentangled representation in an unsupervised manner, by maximizing the mutual information between the observation and a subset of the latent variables. Other works, such as~\cite{lample2017fader,naama}, learn class invariant representation, based on label  supervision which we do not employ, in order to produce disentanglement between domains.  \sab{Unlike our method. these methods are unable to produce the latent representation corresponding the domain specific or domain invariant information}. 

\section{Method}
%\subsection{Problem Setup}
%\label{sec:problem_def}

We transfer content that exists in a sample $b$ in domain $B$ onto a sample $a$ from a similar domain $A$, in which this \textit{domain specific} content is not found. In addition, we also consider the task of weakly supervised semantic segmentation of the domain specific content. That is, given unpaired samples from domains $A$ and $B$, we wish to label (or generate a segmentation mask for) the domain specific part in an image $b\in B$. %Using the same running example, given an image $b \in B$, we seek to label its facial hair.  
We also consider attribute removal. That is, given $b \in B$, we wish to remove the domain specific part of $b$.

Our method consists of five different networks: the common encoder, $E_c$, aims to capture the \textit{common} (or domain invariant) information between domains $A$ and $B$. The separate encoder, $E_s$, aims to capture the \textit{separate} (or domain specific) information in domain $B$. The domain confusion network, $C$, is used to make the encodings generated by $E_c$ for images from both domains indistinguishable. %match the distribution of common encodings from $A$, $\mathbb{P}_{E_c(A)}$, and common encodings from $B$, $\mathbb{P}_{E_c(B)}$, and play an important role the training the aforementioned encoders.
The decoder, $D_A$, generates samples in domain $A$, given a representation that is obtained by the common encoder $E_c$. If that sample comes from domain $B$, the domain-specific content is removed. %no matter which domain they originate from, and is key in feeding information to the next module. For a sample $a \in A$, the decoder is used to reconstruct it. For a sample $b \in B$, the decoder removes the separate content, by using only the common part of its encoding.  

The generation of the image that combines the content of $a$ and the domain specific content of $b$ is done by the decoder $D_B$, which returns two image-sized outputs: $z^\text{raw}$ and $m$. 
%The second decoder $D_B$ is used to generate the domain-B specific component given image $a\in A$ and $b\in B$ and returns two image-sized outputs
\begin{align}
\label{eq:db}
    m(a,b),z^\text{raw}(a,b) = D_B(E_c(a),E_s(b))
\end{align}
where $m(a,b)$ is a soft mask with values between 0 and 1 and $z^\text{raw}$ is an image. It is important to note that the mask and the generated image both depend on the content in $b$ as well as on the image $a$, which determines the placement and other appearance modifications.

\begin{figure}
\centering
%\begin{tabular}{c}
  %\includegraphics[width=0.5\linewidth, clip]{results/glasses_2.jpg} &
%\includegraphics[width=1.0\linewidth, clip]{results/full_diagram.pdf} \\
  %\end{tabular}
    \includegraphics[width=0.8\linewidth, clip]{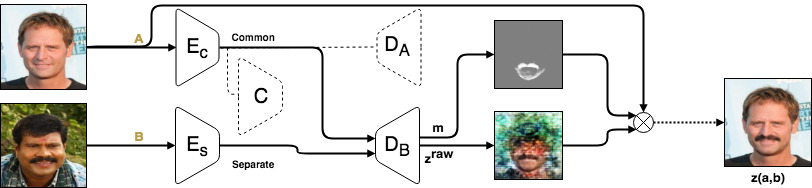}
\caption{An illustration of the inference procedure. The discriminator $C$ and decoder $D_A$ are not used during inference but are included for illustrative purposes. 
%   \includegraphics[width=0.99\linewidth, clip]{results/overviewNlosses.pdf}
% \caption{An illustration of our method. Top: our architecture (yellow box) accepts two images as input, and generates a combination of them, $z$, and a mask, $m$, indicating where the content was transferred spatially. For the segmentation task (blue path), a single image from domain $B$ is fed, and the mask $m$ is the regarded output. Bottom: visualization of the cycle, reconstruction, and regularization losses.
%paths for four of the loss terms. the combination of  part is an illustration of training while bottom part is for inference. (a). Reconstruction for domain $B$. (b) Reconstruction for domain $A$. $D_A$ tries to reconstruct the input. The application of $z$ on the input with no facial hair should give the same input. (c) Cycle consistency loss (d) Weakly supervised segmentation inference. (e). Guided content transfer inference. 
}
  \label{fig:overview}
\end{figure}

%The latter is a soft mask (with values between $0$ and $1$), determining the augmentation location within the image. The former constitutes the respective augmentation content. 
The final output $z$ is a combination of these outputs and image $a$,
\begin{align}
z(a,b) & = m(a,b) \otimes z^\text{raw}(a,b) + (1-m(a,b)) \otimes a,
\end{align}
where $\otimes$ stands for an element wise multiplication.  Fig.~\ref{fig:overview} illustrates the inference step as well as the five networks.% our method as well as the losses employed: 
%We describe the losses employed in our method here:
%The masking module consists of a domain $B$ content generation network, $T$, for samples $b \in B$.
%Given a separate encoding of $b$, the common information of $b$, and the image $b$ itself, $T$ generates a "patch", and a localized position, given by a mask, to which the "patch" is to be placed in $b$. In addition to the mask it is necessary to generate the "patch", as the separate information (e.g the facial hair) may vary slightly when inserted onto a different image. We also note the masking module provides signal for the separate encoder in the disentanglement module, via a reconstruction loss. 

%two main modules, as depicted in Figure \ref{fig:overview}. The first is the disentanglement module, responsible for encoding the domain specific and the domain invariant contents separately, and is inspired by the work of \cite{ori}. The second module, which is the key insight of our proposal, is the masking module. This module locates the region in the target that should be changed, as well as generates the correct patch to be placed there. This allows keeping the unrelated details in tact, facilitating great improvement in generation quality --- as the augmentation focuses on the relevant part of the image, leaving all other details to be taken from the target image directly, without going through the bottleneck of an auto-encoder like module. 

%The disentanglement module consists of five parts: The two encoders perform the disentanglement. 

% In the training phase, a reconstruction loss is used to ensure that, given a sample $b \in B$, the common information, $E_c(b)$ and the separate information, $E_s(B)$ are sufficient sufficient to reconstruct image $b$. In the running example both the facial features and the beard features are needed to reconstruct a face image with beard. 
% Similarly, given image $a \in A$, it should be sufficient to use the common information only, $E_c(b)$ to reconstruct image a. In our running example, only the facial features are needed to reconstruct a face image without beard. 
\paragraph{Domain Confusion Loss.} We seek to ensure that the common encoding, generated by $E_c$, contains only information that is common to both domains. This is done by combining reconstruction losses with a domain confusion loss. The latter employs a discriminator network, $C$, that encourages the encodings of the two domains to statistically match~\cite{ganin2016domain}. %are in
%To do so, we use an domain confusion loss to ensure that the distribution $\mathbb{P}_{E_c(A)}$ of $E_c(A)$ matches the distribution $\mathbb{P}_{E_c(B)}$ of $E_c(B)$. Specifically:
% MAKE SURE CORRECT DEFINITIONS HERE
%\begin{small}
\begin{align}
\label{eq:3}
\mathcal{L}_{DC} &:= \frac{1}{|S_A|} \sum_{a\in S_A} l(C(E_c(a)),1)+ \frac{1}{|S_B|}\sum_{b\in S_B} l(C(E_c(b)),1)
\end{align}
%\end{small}
where $S_A$ and $S_B$ are the training sets sampled from the two domains and $l(p,q) = -(q\log(p)+(1-q)\log(1-p))$ is the binary cross entropy loss for $p\in\left[0,1\right]$ and $q\in\{0, 1\}$.%\ab{this is over complicated, why not just write $log(C(E_c(b)))$ and mention that it's a cross-entropy loss? If we do keep it, what does $q\in\{0, 1\}$ mean?}

Our formulation of the domain confusion loss is similar to that of \cite{tzeng} except where for $\mathcal{L}_{DC}$, $E_c$ attempts to fool the discriminator $C$, so that the encodings of both domain A and domain B would be classified as $1$. Namely, $C$ tries to distinguish between encodings of domain A and B, while $E_c$ attempts to produce an encoding which is indistinguishable for $C$.

While $E_c$ attempts to make the two distributions indistinguishable, $C$ is trained in an adversarial manner to minimize the following objective: 
%\begin{small}
\begin{align}
    \mathcal{L}_{C} &:= \frac{1}{|S_A|} \sum_{a\in S_A} l(C(E_c(a)),0)+ \frac{1}{|S_B|}\sum_{b\in S_B} l(C(E_c(b)),1)
\end{align}
%\end{small}

\paragraph{Reconstruction Loss}
\label{sec:reconstuction}

%\subsubsection{Reconstruction of samples in $A$}

The domain confusion loss ensures that the common encoder, $E_c$, does not encode any separate information from domain $B$. %However, we need to also ensure that (i) $E_c$ captures all the common information between $A$ and $B$, but not less, and (ii) $E_s$ captures all the information in $B$ which is not in $A$, but not less.
%\end{enumerate}
For samples $a \in A$, we also need to verify that the information in $E_c(a)$ is sufficient to reconstruct it, ensuring that all the information of domain $A$ is encoded by $E_c$. We use
\begin{align}
\label{eq:5}
\mathcal{L}_{Recon1}^A &:= \frac{1}{|S_A|} \sum_{a\in S_a} \| D_A(E_c(a)) - a \|_{ 1}
\end{align}
where $\|\|_{1}$ is the L1 loss directly applied to the RGB image values. 

Similarly, we wish to verify that the information encoded by $E_s$ is sufficient for reconstructing the separate details, so that $E_s(B)$ contains the domain specific information of domain $B$. Given an image $b\in B$, we do this by removing the separate information from it, using $D_A(E_c(b))$, and adding it back:
%The reconstruction loss terms for domain $B$ take the form
\begin{align}
\label{eq:recon_b} \mathcal{L}_{Recon1}^B &:= \frac{1}{|S_B|}\sum_{b\in S_B} \| z'(D_A(E_c(b)),b,b) - b \|_{ 1},
\end{align}
where $z'$ is defined as:
\begin{align}
\label{eq:7}
z'(c,a,b) = m(a,b) \otimes z^\text{raw}(a,b) + (1-m(a,b)) \otimes c
\end{align}
For Eq.~\ref{eq:recon_b}, we use $z'$ instead of $z$. This is so $E_c$ is not applied on $D_A(E_c(b))$, but directly on $b$. In both cases, one recovers the common information of $b$, but when using $z$, additional error is introduced thought the use of $D_A \circ E_c$.

Finally, we reinforce the roles of the two domains by encouraging the mask to be minimal. In our experiments, we saw that explicitly penalizing the mask size, or using other traditional regularization terms, yielded inferior results, as shown in Sec.~\ref{sec:ablation}. Instead, we achieve this goal in a softer way, by running samples from each domains through both inputs of our transfer pipeline and favouring successful reconstruction: 
%In this loss, we reconstruct $b$ twice, using the shared information that is extracted (by $E_c$ of Eq.~\ref{eq:db}), either from the image $b$ itself or from its version in which the domain specific part is first removed and then added back. Running the samples of domain $a$ through the reconstruction pipeline of domain $B$ reinforces the roles of the two domains, by encouraging the mask in this case to be minimal:
\begin{align}
\label{eq:recon_ab}  \mathcal{L}_{Recon2}^A := \frac{1}{|S_A|}\sum_{a\in S_A}  \|z(a,a) - a \|_{ 1} \text{ ~~~~~~~~~~~~~~~~ } 
 \mathcal{L}_{Recon2}^B := \frac{1}{|S_B|}\sum_{b\in S_B}  \|z(b,b) - b \|_{ 1} 
\end{align}

The first term of the loss introduced in Eq.~\ref{eq:recon_ab} ($\mathcal{L}_{Recon2}^A$) encourages a minimal distance between $z(a,a)$ and $a$, where
 $z(a,a)= z^{raw}(a,a) \otimes m(a, a) + a\otimes(1-m(a,a))$. Ideally, $z^{raw}$ would be equal to $a$, but since we use an encoder and a decoder which cannot auto-encode perfectly, we get that there is some distance between $z^{raw}$ and $a$. Hence, in order to minimize the distance between $z(a,a)$ and $a$, the network minimizes the size of the mask. Similar argument holds for  $\mathcal{L}_{Recon2}^B$.

\paragraph{Cycle Consistency Losses}
\label{sec:cycleloss}
Cycle consistency in the latent spaces is used as an additional constraint to encourage disentanglement. Specifically, we have:
\begin{align}
\mathcal{L}_{Cycle} &:= \frac{1}{|S_A||S_B|} \sum_{a\in S_A, b\in S_B}  \| E_c(z(a, b)) - E_c(a) \|_2 + \|E_s(z(a, b)) - E_s(b) \|_{2} 
%\mathcal{L}_{CycleSep} &:= \frac{1}{|S_A||S_B|}  \sum_{a\in S_A, b\in S_B} \| E_s(z(a, b) - E_s(b) \|_2
\end{align}
where $\|\|_2$ is the MSE loss.

The overall loss term we minimize is:
\begin{align}
\mathcal{L} & = \mathcal{L}_{DC} + \lambda_1\mathcal{L}_{Recon1}^A + \lambda_2\mathcal{L}_{Recon1}^B  + \lambda_3\mathcal{L}_{Cycle} + \lambda_4\mathcal{L}_{Recon2}^{A}
+ \lambda_5\mathcal{L}_{Recon2}^B\nonumber
\end{align}
where $\lambda_1, \dots, \lambda_5$ are positive constants. We train a discriminator $C$ separately to minimize $L_C$.

\noindent{\bf Inference~~}
\label{sec:inference}
The network's architecture is provided in appendix~\ref{sec:arc}.  Once trained, the networks can be used for unsupervised content transfer and weakly supervised segmentation of the domain specific information. In the first case, we generate examples $z(a,b)$ for $a\in A,b\in B$. In the second, we consider the mask generated by feeding an image $b$ from domain $B$ to \textbf{both} inputs $m(b,b)$, then apply a threshold to get a binary mask. As shown in appendix Fig.~\ref{fig:threshold} the method is largely insensitive to the exact value of the threshold.  

The network can also be used for attribute removal by generating $z_{unmasked} := D_A(E_c(b))$.
$z_{unmasked}$ is $b$ with its separate part removed. In order to avoid missing reconstructed facial features the generated output is calculated as:
\begin{align*}
m(b,b), z &:= D_B (E_c(b), E_s(b)) \\
z_{removed} &:= (1-M(b,b)) \otimes b + M(b,b) \otimes z_{unmasked}. 
\end{align*} 
where M is the binarized mask of the soft mask m.

\section{Experiments}

We evaluate our method for guided content transfer, out of domain manipulation, attribute removal, sequential content transfer, sequential attribute removal and content addition, and weakly supervised segmentation of the domain specific content.

\noindent{\bf Guided Content Transfer\quad} We employ three attributes that are expressed locally in the images of the celebA dataset~\cite{celeba}: smile, facial hair, and glasses. In each case, we consider $B$ to be the domain of images with the attribute, and $A$ to be the domain without it. 
%We used a $90\%$ train, $10\%$ test split.

\begin{figure}
\centering
%\begin{tabular}{c}
  %\includegraphics[width=0.5\linewidth, clip]{results/glasses_2.jpg} &
  \includegraphics[width=0.8\linewidth, clip]{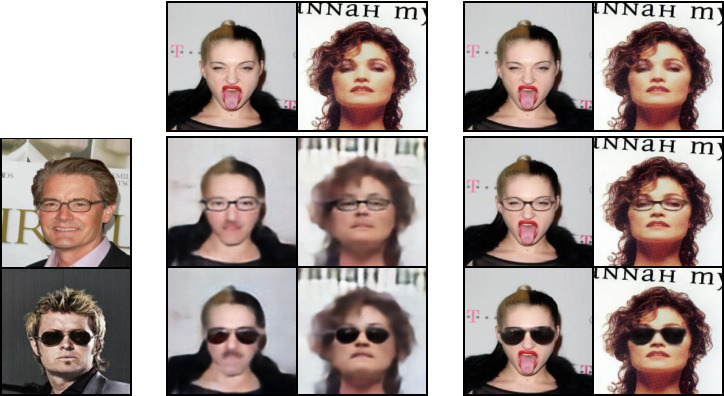} \\
  %\hspace{-1.0cm} (a)   ~~~~~~~~~~~~~~~~~~~~~~~~~~~~~~~~~~~~      (b) ~~~~~~~~~~~~~~~~~~~~~~~~~~~~~~~~~~~~~~~~~~~~~~~~~ (c)
  %\end{tabular}
\caption{Glasses from guide images in domain $B$ (left) augment the glasses-less source images from domain $A$ (top). The content transfer of \cite{ori} (middle) is compared to our results (right). 
%: the  top row is the source images in domain $A$. The others incorporate the glasses from the corresponding row of (a). (c) The same mapping for the method of \cite{ori}
}
  \label{fig:glasses_comapre}
\end{figure}

We first consider the ability to add the separate part of an image $b \in B$ to the common part of $a \in A$. This is shown for the domain of glasses, in Fig.~\ref{fig:glasses_comapre},  compared to the baseline method of \cite{ori}. As can be seen, only the local structure of the glasses is changed, whereas in the baseline many low level details are lost (for example, the background writing) and unnecessary changes are made (for example, an open mouth is replaced with a closed one, or facial hair is added, changing the identity of the source image). 
%Additional results for for glasses and for the remaining attributes are shown in appendix due to lack of space.
Furthermore, Fig.~\ref{fig:orientation} demonstrates the ability of our method to accommodate for different orientations of the source image $a$, and to properly adapt the glasses from $b$ to the correct orientation. Please refer to the appendix~\ref{sec:add_res} for more examples.

To assess the quality of the domain translation, we conduct a handful of quantitative evaluations. In Tab.~\ref{tab:kid_fid}, we consider the Frechet Inception Distance (FID) \cite{fid} and Kernel Inception Distance (KID) \cite{kid} scores of images with the common part of $a$ and separate part of $b$ over a test set of images from domains $A$ and $B$. The FID score is a commonly used metric to evaluate the quality and diversity of produced images; KID is a recently proposed alternative for FID. We note that these values should only be used comparatively, as the size of the test set used affects the score magnitude. %may vary from other papers. 
As can be seen, our method scores significantly better. %compared to the baseline method 

% Image quality - content transfer for facial hair attribute: 185 images, 10 subsets of 150 images compared to 8000 real images of facial hair. Glasses: 66 images. In KID we used $\gamma = 0.01$. }

\begin{table}[t]
%\begin{small}
\begin{minipage}[c]{0.5940\linewidth}
\caption{FID and KID scores (lower is better) for generated images using the common part of $a \in A$ and the separate part of $b \in B$. As real images, we consider the images in $A$. For KID we used $\gamma = 0.01$, kernel $k(x,y) = (\gamma x^T y + 1)^3$.}
%\vspace{-1.2mm}
  \label{tab:kid_fid}
%\begin{center}
  \begin{tabular}{@{}l@{~}c@{~}c@{~}c@{~}c@{}}
    \toprule
    & \multicolumn{2}{c}{Facial hair} & \multicolumn{2}{c}{Glasses}\\
    \cmidrule(lr){2-3}
    \cmidrule(lr){4-5}
  	Method & FID  	  &  KID  & FID 	  &  KID  \\ %
  	\midrule
    Real images & 85.4$\pm$2.9 & 2.5$\pm$0.2 & 115.5$\pm$3.8 & 0.1$\pm$0.3 \\  	Ours & 90.7$\pm$1.8 &  3.5$\pm$0.1 & 134.9$\pm$4.8 & 5.2$\pm$0.8 \\
  	Press et al. & 139.4$\pm$ 1.9 & 16.8$\pm$0.5 & 178.5$\pm$3.2 &  14.6$\pm$1.2 \\
    \bottomrule
\end{tabular}
\end{minipage}%
\hfill
\begin{minipage}[c]{0.38\linewidth}
\vspace{3mm}
%\end{center}
  \caption{The accuracy of generated images according to a pretrained classifier distinguishing between $A$ and $B$.}%samples in $A$ and samples in $B$. We sample $a \in A$, $b \in B$ from the test samples and consider the ratio of predicted label $B$ on the generated images of each method.}
  \label{tab:classifier}
  \begin{tabular}{@{}l@{~}c@{~}c@{~}c@{}}
    \toprule
     &  Smile & Glasses & Beard   \\ 
     %&   &  & Hair   \\ 
    \midrule
  	Fader & 93.9 \% & 93.6\% & 81.8\% \\ 
    Press et al. &  98.9\% & 94.8\% & 88.1\% \\
    MUNIT & 8.5\% & 8.3\% & 7.2\%\\
    DRIT & 9.2\% & 7.4\% & 6.5\%\\
    Ours & 99.2\%	 & 96.2\% & 88.0\% \\ %
    \bottomrule
\end{tabular} 
%\smallskip
\end{minipage} \\
\vspace{-0.3cm}
\caption{An evaluation of the cosine similarity (higher is better) before and after translation between the VGG-face descriptors. Shown are average results over 100 random images created by sampling $a$ and $b$ from the test sets.}
\label{tab:similarity}
\smallskip
\centering
  \begin{tabular}{@{}l@{~~~}c@{~}c@{~~~}c@{~}c@{~}c@{}}
    \toprule
    & \multicolumn{3}{c}{A to B mapping} & \multicolumn{2}{c}{Transfer A' to B'}\\
    \cmidrule(lr){2-4}
    \cmidrule(lr){5-6}
    &  Facial hair & Glasses & Smile & Facial hair  & Glasses  \\     
    & male to male & all genders & all genders & male to female %$A',B'$  
    & train women, test men\\
    \midrule
    Ours &  0.89 &  0.84 & 0.94  & 0.90 & 0.82 \\
    \cite{ori} & 0.73 & 0.68 & 0.73 & 0.64 & 0.59\\ 
    %Euclidean Distance (Ours) & 59 & 76 & 41 & 69 & 74\\
    %Euclidean Distance (\cite{ori}) & 94 & 109 & 95 & 124 & 115\\
    \bottomrule
\end{tabular}
%\smallskip
  \caption{User study (questions (1), (2) and (3)) showing preference to our method vs.~\cite{ori}, see text.}% for details. }
    \label{tab:user_study}
    \smallskip
\centering
  \begin{tabular}{@{~}lccccccccc@{~}}
    \toprule
        & \multicolumn{6}{c}{A to B mapping} & & \multicolumn{2}{c}{A' to B' shift}\\
    \cmidrule(lr){2-8}
    \cmidrule(lr){9-10}
    &  Facial & Glasses & Smile & Hand- & Two & Remove & Facial & Facial  & Glasses  \\    
    &  hair & all & all & bags & Attrs & Smile & hair & hair  & train women \\     
    & male to  & genders & genders &  &  & Add & swap & female  & test men\\
    & male &  & &  &  & Glasses & & $A',B'$  & \\
    \midrule
    (1) &  96\% &  95\% & 55\% & 87\% & 91\% & 83\% & 93\% & 91\% & 93\% \\ %
    (2) & 84\% & 82\% & 43\% & 72\%& 93\%& 90\%& 83\% & 70\% & 86\%\\ 
    (3) & 97\% & 95 \% & 95\% & 90\%  & 95\%& 91\% & 91\% & 91\% & 97\%\\
    \bottomrule
\end{tabular}
%\end{small}
\end{table}

We also consider the ability of our method to transfer the separate part of $b$ to the target image. To do so, we use a pretrained classifier to distinguish between domains $A$ and $B$ (on the respective training sets) and consider the score of the translated images\ab{.}
%, evaluating against membership in $B$. 
These results are reported in Tab.~\ref{tab:classifier}, and show a clear advantage to our method. 
As expected, the MUNIT and DRIT methods~\cite{munit,Lee_2018_ECCV} are not competitive in this metric, since they transfer style and not content. Additionally, in contrast to our method, Fader networks (\cite{lample2017fader}) transfer to $B$ without the use of a specific guide image $b$ from this domain.
%As expected, the methods of~\cite{munit,Lee_2018_ECCV} transfer style and not content and are not competitive in this metric. Fader networks~\cite{lample2017fader} transfer to $B$ without the ability to use a specific guide image $b$ from this domain.

% \begin{table}[t]
% \centering
%   \caption{The accuracy on generated images of a pretrained classifier that distinguishes between samples in $A$ and samples in $B$. We sample $a \in A$, $b \in B$ from the test samples and consider the ratio of predicted label $B$ on the generated images of each method.}
%   \label{tab:classifier}
%   \begin{tabular}{lccc}
%     \toprule
%      &  Smile & Glasses & Beard   \\ 
%     \midrule
%   	Fader ~\cite{lample2017fader} & 93.9 \% & 93.6\% & 81.8\% \\ 
%     \cite{ori} &  98.9\% & 94.8\% & 88.1\% \\
%     MUNIT~\cite{munit} & 8.5\% & 8.3\% & 7.2\%\\
%     DRIT~\cite{Lee_2018_ECCV} & 9.2\% & 7.4\% & 6.5\%\\
%     Ours & 99.2\%	 & 96.2\% & 88.0\% \\ %
%     \bottomrule
% \end{tabular}

% \end{table}

To evaluate if the source identity is preserved, we compute 
%Our ability to preserve the identity of the source images is evaluated by computing 
the cosine similarity of the pretrained VGG-face network~\cite{vggface2}. High values indicate preserved identity. 
%As can be seen from 
Tab.~\ref{tab:similarity} indicates that our results exhibit a much better similarity to source images than baseline methods. 
%those of the baseline method.%. This is consistent with the visual comparison, where low level details are preserved.

\begin{figure}
\centering
\begin{tabular}{c}
  \includegraphics[width=0.85\linewidth, clip]{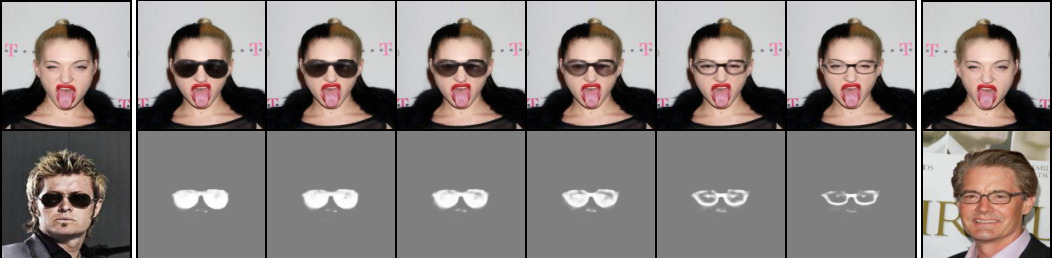}
  \end{tabular}
\caption{\ab{Interpolation} between $E_s(b_1)$ \ab{(bottom left)} and $E_s(b_2)$ \ab{(bottom right)} for $ b_1,b_2\in B$, while fixing the source image $a \in A$ (top ends). The generated images \ab{(top)} and masks \ab{(bottom)} are shown. }
  \label{fig:interpolation}
  \vspace{-0.7cm}
\end{figure}

To evaluate the interpretability of the latent space,  we interpolate between the latent code of the separate parts of images $b_1 \in B$ and $b_2 \in B$ with the common latent code of an image $a \in A$. This is shown in Fig.~\ref{fig:interpolation} and appendix ~\ref{sec:transfer}. 
Note the mask changes throughout the interpolation.

% \begin{table*}[t]
% \caption{An evaluation of the cosine similarity (higher is better) before and after translation between the VGG-face descriptors. Shown are average results over 100 random images created by sampling $a$ and $b$ from the test sets.}
% \centering
%   \begin{tabular}{lccccc}
%     \toprule
%     & \multicolumn{3}{c}{A to B mapping} & \multicolumn{2}{c}{Transfer A' to B'}\\
%     \cmidrule(lr){2-4}
%     \cmidrule(lr){5-6}
%     &  Facial hair & Glasses & Smile & Facial hair  & Glasses  \\     
%     & male to male & all genders & all genders & female $A',B'$  & train women, test men\\
%     \midrule
%     Ours &  0.89 &  0.84 & 0.94  & 0.90 & 0.82 \\
%     \cite{ori} & 0.73 & 0.68 & 0.73 & 0.64 & 0.59\\ 
%     %Euclidean Distance (Ours) & 59 & 76 & 41 & 69 & 74\\
%     %Euclidean Distance (\cite{ori}) & 94 & 109 & 95 & 124 & 115\\
%     \bottomrule
% \end{tabular}
% \smallskip
% \smallskip
%   \label{tab:similarity}
% \end{table*}

\noindent{\bf User study\quad} To further strengthen the evaluation, we conduct a user study. We randomly sample $20$ images from $a \in A$ and $b \in B$ and consider the translated image of our method vs. that of \cite{ori}. We conduct three experiments where the user is asked to select: (1) the translated image that matches the distribution of $B$ more closely, (2) Given the guide image $b$, in which translated image, the separated part of $b$ is better transferred, and (3) Given the source image $a$, which translated image better preserves the facial features of $a$. Average scores are reported in Tab.~\ref{tab:user_study}. For the tasks of facial hair and glasses, we score consistently higher than the baseline method. For smile, our ability to produce realistic smiling faces is slightly higher, the ability to transfer the smile from the source image is slightly worse, while our ability to preserve the identity of the source image is significantly higher. This probably stems from the smile taking place not only in the specific mouth region. 

\begin{figure}[t]
  \centering
  \begin{minipage}[c]{0.49\linewidth}
  \vspace{0.9cm}
  \begin{tabular}{c}
  \includegraphics[width=0.899\linewidth, clip]{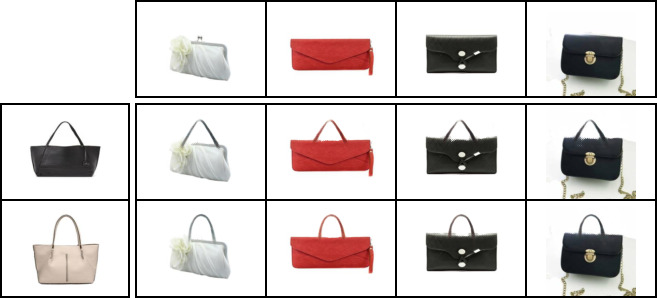}
  \end{tabular}
\captionof{figure}{Adding a handle to a handbag.}
  \label{fig:handbags}

  \end{minipage}%
\hfill
\begin{minipage}[c]{0.49\linewidth}
\captionof{table}{Mean and SD IoU for the two hair segmentation benchmarks.}
  \label{tab:segmentation_numerics}
\vspace{0.5cm}

  \begin{tabular}{lcc}
    \toprule
  	Method & Women's hair	  &  Men's hair \\ %
  	\midrule
  	Ours & 0.77$\pm$ 0.15 & 0.77 $\pm$ 0.13 \\
  	Press et al. & 0.67$\pm$ 0.13 & 0.58 $\pm$ 0.11 \\ 
    Ahn \& Kwak. & 0.54$\pm$ 0.10 & 0.52 $\pm$ 0.10 \\
    CAM & 0.43$\pm$ 0.09 & 0.56 $\pm$ 0.07 \\
    \bottomrule
\end{tabular}
\end{minipage}
\end{figure}

\noindent{\bf Out of domain manipulations\quad}
We also consider the ability of the learned model handle a domain shift, i.e. to perform a translation from a domain which was not seen during training. For example, we train on female faces without glasses as domain $A$, and female faces with glasses for domain $B$. At test time, $A$ is replaced with a domain $A'$ of male faces, and we are asked to transfer the glasses onto the male's face, generating a domain $B'$ from which we see no train or test samples. Quantitative evaluation is provided in Tab.~\ref{tab:similarity}, showing a negligible difference in quality for our method, and a significant one for the baseline method. Visual results can be found in appendix~\ref{sec:transfer}, where we also consider out of domain LFW  dataset~\cite{lfw} as well as extremely out of domain images, which our method successfully handles. 
\paragraph{Handbags} We also consider the domain of handbags~\cite{handbag}, where we split this domain into images with a handle ($B$) and those without ($A$). The transfer results are illustrated in Fig.~\ref{fig:handbags}. The generated mask and raw outputs are clearly adapted to the bag on which the handle content is placed. The user study in Tab.~\ref{tab:user_study} evaluates these results. Please refer to appendix~\ref{sec:handbags} for visual comparison.

% \begin{figure}
% \centering
% \begin{tabular}{c}
%   %\includegraphics[width=0.5\linewidth, clip]{results/glasses_2.jpg} &
%   \includegraphics[width=0.99\linewidth, clip]{results/bags_grid_v2.jpg}
%   \end{tabular}
% \caption{Translation from handbags without a handle to \ab{ones} with a handle. }
%   \label{fig:handbags}
% \end{figure}

\noindent{\bf Attribute removal\quad} While our method is more general than attribute transfer methods,
%such as \cite{lample2017fader, attgan, stgan}, 
it can be used to remove a given attribute as shown in Fig~\ref{fig:removal_text_new}; see  appendix~\ref{sec:removal} for a full qualitative comparison to the literature methods. A quantitative evaluation is given in Tab~\ref{tab:removaltab}. For generation quality we use KID and FID; for successful attribute removal, a pretrained classifier is used to measure the percentage of test images without the attribute, and for similarity with the source image, a perceptual loss is used (using the features of a VGG-face~\cite{vggface2} network). Our method is significantly superior in terms of generation quality over all baseline methods for all tasks and presents a good tradeoff between fidelity and transformation success. 
For facial hair removal, \cite{ori} and \cite{lample2017fader} are superior in terms of classifier accuracy, yet their generation quality is far worse (blurry images) and the similarity to the source image is significantly impaired. As the comparison images in appendix~\ref{sec:removal} show,  \cite{lample2017fader} achieves higher accuracy by producing female images while \cite{ori} makes the persons younger looking. \cite{attgan} has slightly superior similarity score, but is worse on removing the facial hair and has worse generation quality. For smile, \cite{stgan} is slightly superior in removing the smile in terms of accuracy, yet worse on generation quality and similarity to the source image. For glasses, all classifier scores are close to 100\% meaning an almost perfect glasses removal, yet our similarity score and generation quality is higher.

\begin{figure}[t]
  \centering
  \begin{minipage}[c]{0.26\linewidth}
  \captionof{figure}{Attr removal.\label{fig:removal_text_new}}
  \begin{tabular}{c}
  \includegraphics[width=0.78\linewidth, clip]{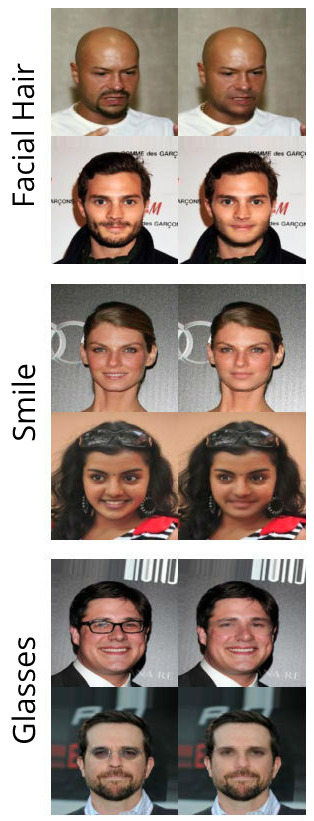}
  \end{tabular}

  \end{minipage}%
\hfill
\begin{minipage}[c]{0.72\linewidth}
\centering
  \captionof{table}{Attribute removal for the task of Smile, Facial hair and Glasses. }
    \label{tab:removaltab}
    \smallskip
\begin{tabular}{llcccc}
\toprule
Task        & Method & KID   & FID   & Class. & Sim.   \\ 
\midrule
Smile       & Ours   & 2.6 $\pm$ 0.4  & 120.0 $\pm$ 2.6 & 96.9\% & 0.96\\
            & Press et al. &  15.0 $\pm$ 0.6 & 167.7 $\pm$ 0.3 & 96.9\% & 0.81 \\
            & He et al. & 4.1 $\pm$ 0.4  & 127.7 $\pm$ 4.5 & 96.9\% & 0.95\\
            & Liu et al.  & 4.3 $\pm$ 0.3 & 129.0 $\pm$ 3 & 98.4\% & 0.92\\
            & Fader & 11.3 $\pm$ 0.7 & 155.6 $\pm$ 4.7 & 93.7 \% & 0.89\\
            \midrule
Mustache & Ours   & 1.9 $\pm$ 0.5 & 119.0 $\pm$ 0.8 & 95.3 \% & 0.95\\
            & Press et al. &  16.6 $\pm$0.8 & 175.9 $\pm$ 1.4 & 100.0\% &  0.80\\
            & He et al. & 4.6  $\pm$ 0.5 & 130.0 $\pm$ 3.0 & 87.5\% &  0.96\\
            & Liu et al.  & 14.0 $\pm$ 0.6 & 160.0 $\pm$ 3.3 & 87.5\% &  0.85\\
            & Fader & 14.1 $\pm$ 0.6 & 162.6 $\pm$ 1.5 & 98.4  \% &  0.76\\
            \midrule
Glasses     & Ours   & 5.2$\pm$ 0.5 & 136.5$\pm$ 2.6 & 99.2\% &  0.87\\
            & Press et al. & 15.3$\pm$ 0.5 &172.0 $\pm$ 4.7 & 100.0\% & 0.73\\
            & He et al. & 8.3 $\pm$ 0.9 & 141.4$\pm$6.8 & 100.0\%  & 0.84\\
            & Liu et al.  & 6.8 $\pm$ 0.3 & 141.8 $\pm$ 4.8 & 98.4\%  & 0.86\\
            & Fader & 12.5$\pm$ 0.3 & 137.7$\pm$ 4.2 & 100.0\% & 0.76\\
            \bottomrule
\end{tabular}
\end{minipage}
%\end{figure}
%\begin{figure}
~\\
\centering
\label{fig:removeadd}
\vspace{-0.1cm}
\includegraphics[width=0.7\linewidth,clip]{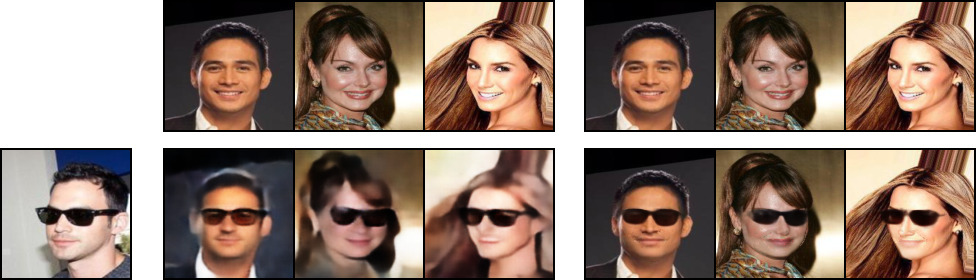} 
\caption{Removal of smile and addition of glasses according to the guided image on the left. In the middle, the translation of \cite{ori} and on the right, our result.}
  
\end{figure}

% \begin{table*}[t]
% \centering
%   \caption{Attribute removal for the task of Smile, Facial hair and Glasses. }
%     \label{tab:removaltab}
%     \smallskip
% \begin{tabular}{llcccc}
% \toprule
% Task        & Method & KID   & FID   & Class. & Sim.   \\ 
% \midrule
% Smile       & Ours   & 2.6 $\pm$ 0.4  & 120.0 $\pm$ 2.6 & 96.9\% & 0.96\\
%             & Press et al. &  15.0 $\pm$ 0.6 & 167.7 $\pm$ 0.3 & 96.9\% & 0.81 \\
%             & He et al. & 4.1 $\pm$ 0.4  & 127.7 $\pm$ 4.5 & 96.9\% & 0.95\\
%             & Liu et al.  & 4.3 $\pm$ 0.3 & 129.0 $\pm$ 3 & 98.4\% & 0.92\\
%             & Fader & 11.3 $\pm$ 0.7 & 155.6 $\pm$ 4.7 & 93.7 \% & 0.89\\
%             \midrule
% Facial & Ours   & 1.9 $\pm$ 0.5 & 119.0 $\pm$ 0.8 & 95.3 \% & 0.95\\
% hair            & Press et al. &  16.6 $\pm$0.8 & 175.9 $\pm$ 1.4 & 100.0\% &  0.80\\
%             & He et al. & 4.6  $\pm$ 0.5 & 130.0 $\pm$ 3.0 & 87.5\% &  0.96\\
%             & Liu et al.  & 14.0 $\pm$ 0.6 & 160.0 $\pm$ 3.3 & 87.5\% &  0.85\\
%             & Fader & 14.1 $\pm$ 0.6 & 162.6 $\pm$ 1.5 & 98.4  \% &  0.76\\
%             \midrule
% Glasses     & Ours   & 5.2$\pm$ 0.5 & 136.5$\pm$ 2.6 & 99.2\% &  0.87\\
%             & Press et al. & 15.3$\pm$ 0.5 &172.0 $\pm$ 4.7 & 100.0\% & 0.73\\
%             & He et al. & 8.3 $\pm$ 0.9 & 141.4$\pm$6.8 & 100.0\%  & 0.84\\
%             & Liu et al.  & 6.8 $\pm$ 0.3 & 141.8 $\pm$ 4.8 & 98.4\%  & 0.86\\
%             & Fader & 12.5$\pm$ 0.3 & 137.7$\pm$ 4.2 & 100.0\% & 0.76\\
%             \bottomrule
% \end{tabular}
% \end{table*}

\noindent{\bf Sequential content transfer\quad} 
Our method enables a sequential addition of guided content from different guide images and from different domains by applying our method sequentially. Fig~\ref{fig:two_att_transfer} considers this case for glasses and facial hair addition. Our method significantly outperforms \cite{ori}, as it does not wastefully reconstruct the facial features twice as shown in Tab~\ref{tab:user_study} (``two attributes'') for adding facial hair and glasses.

\noindent{\bf Attribute removal and content addition\quad} Given the ability of our method to remove a specific attribute, one can perform guided content transfer between any given domain, A and B, each with it separate domain specific information. First, we remove the domain specific attribute of domain A and then perform guided content addition for domain B. For example, in Fig.~\ref{fig:removeadd} smile is removed and glasses are then added, see  appendix~\ref{sec:two_att} for more results, as well as facial hair swap in appendix Fig~\ref{fig:replace_mustache}. As for sequential content transfer, we do not wastefully reconstruct the facial features and so significantly outperform \cite{ori} as can be seen in Tab~\ref{tab:user_study} for the task of smile removal and glasses addition as well as facial hair swap (removing and adding facial hair). 

\noindent{\bf Weakly supervised segmentation\quad} We consider the task of segmenting women's and men's hair. For men, $A$ consists of bald men, while $B$ contains men with dark hair. For women, $A$ consists of women with blond hair, while $B$ contains women with black hair. We evaluate our method using the labels given in Borza et al.~\cite{hairlabels}. 

We generate the segmentation using the method described in Sec.~\ref{sec:inference}. We compare our method to \cite{ori}, where we take the translated image and subtract, in pixel space, the source image \ab{from it}. We also compare to the results obtained by the recent weakly supervised segmentation method of Ahn and Kwak~\cite{app8}, which performs segmentation using the same level of supervision we employ, \ab{using published code}. In addition, we compare to CAM \cite{zhou2016learning}, where we train an  Inception-V3 network to classify between the domain and extract localization from the classifier, which we then binarize to get a segmentation mask.

\begin{figure}[t]
  \centering
  \begin{minipage}[c]{0.49\linewidth}
  \begin{tabular}{l}
    \includegraphics[width=0.95\linewidth]{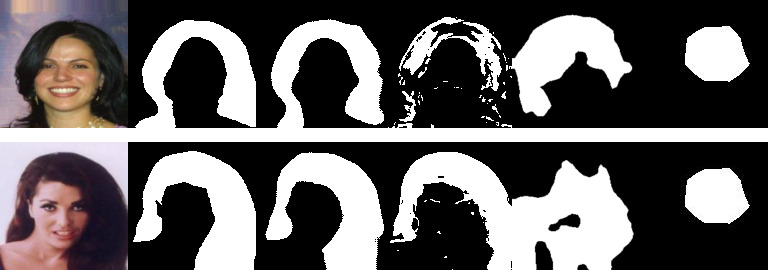}\\%&    \includegraphics[trim=0 0 0 1670, clip,width=0.48\linewidth]{results/women_segmentation.jpg} \\
   ~~~ (a)   ~~~~~~      (b) ~~~~~~ (c) ~~~~~~ (d) ~~~~~~~(e) ~~~~~~~(f)~~~~~\\% & ~~~~~~~ (a)   ~~~~~~~~~      (b) ~~~~~~~~ (c) ~~~~~~~ (d) ~~~~~~~ (e) 
  \end{tabular}
    \captionof{figure}{Segmentation of women's hair. (a) original image, (b) ground truth segmentation, (c) our results, (d) the results of~\cite{ori}, (e) the results of~\cite{app8}, (f) the results of CAM.}
    \label{fig:segmentation_woman}
  \end{minipage}%
\hfill
\begin{minipage}[c]{0.49\linewidth}
\captionof{table}{The effect of removing losses. Shown are classifier accuracy, cosine similarity, KID, and percentage of mask from the total size of the face.}
  \label{tab:ablation_numerics}
%\begin{center}
\smallskip
  \begin{tabular}{@{}lc@{~~}c@{~~}c@{~~}c@{}}
    \toprule
  	& Class. & Sim. &  KID  & Perc. \\ %& FID \\ %
  	\midrule
  	$\mathcal{L}$ & 88.1\% & $0.89$ & $3.5\pm0.1$ & 23\%  \\ % $90.7\pm1.8$ \\
  	w/o $\mathcal{L}^A_{Recon2}$ & 88.5\% & $0.85$ &  $4.1\pm0.5$ & 34\%  \\ % $91.5\pm0.7$ \\
  	w/o $\mathcal{L}^B_{Recon2}$ & 88.1\%& $0.87$ & $4.2\pm0.4$ & 65\%  \\ % $90.4\pm1.9$\\
  	w/o $\mathcal{L}_{Cycle}$ &  67.1\% & $0.95$ & $4.1 \pm 0.9$ & 29\% \\ %& $95.1 \pm 1.1$\\
  	w/o $\mathcal{L}^B_{Recon1}$  & 9.4\% & $1.0$ & $4.3 \pm 0.7$ & 0\% \\ %& $94.4 \pm 1.5$\\
  	w/o $\mathcal{L}^A_{Recon1}$ & 9.7\% & $1.0$ & $4.6 \pm1.0$ & 0\%\\ %& $96.9 \pm 3.9$\\
  	  	w/o $\mathcal{L}_{DC}$ & 9.5\% & $1.0$ & $5.0\pm 1.0$ & 0\%\\ %& $96.9 \pm 3.9$\\
    \midrule
    L2 reg & 88.0\% & $0.82$ & $4.6 \pm 0.7$ & 33\%  \\
  %  \midrule
    L2 recon \#1 & 87.7\% & $0.89$ & $3.3 \pm 0.5$ & 22\%  \\
    L2 recon \#2 & 74.2\% & $0.93$ & $4.2 \pm 0.6$ & 30\%  \\
    \bottomrule
\end{tabular}
\end{minipage}
\end{figure}

As can be seen in Fig.~\ref{fig:segmentation_woman}, our results provide smooth labeling of the hair, while \cite{ori} yield a \ab{broken one with} unnecessary details. The result of~\cite{app8} \ab{also lacks in comparison}. CAM is unable to generate the required shape, as the classifier always focuses on the same place.
%is worse than ours at capturing the structure of the hair. 
Similar results are shown for man's hair in the appendix~\ref{sec:segment}. Our results are also superior \ab{quantitatively}, as shown in Tab.~\ref{tab:segmentation_numerics} for the Intersection over Union (IoU) measure. We also perform semantic segmentation for both glasses and facial hair, refer to appendix~\ref{sec:segment}.
The success of our method stems from requiring the network to minimally add the separate content in the correct location to reconstruct $b\in B$, thus localizing the separate content. 
% \begin{table}[t]
% \caption{Mean and SD IoU for the two hair segmentation benchmarks.}
%   \label{tab:segmentation_numerics}
% \begin{center}
%   \begin{tabular}{lcc}
%     \toprule
%   	Method & Women's hair	  &  Men's hair \\ %
%   	\midrule
%   	Ours & 0.77$\pm$ 0.15 & 0.77 $\pm$ 0.13 \\
%   	\cite{ori} & 0.67$\pm$ 0.13 & 0.58 $\pm$ 0.11 \\ 
%     Ahn et al.~\cite{app8} & 0.54$\pm$ 0.10 & 0.52 $\pm$ 0.10 \\
%     \bottomrule
% \end{tabular}
% \end{center}
% \end{table}

% \begin{figure}[t]
%     \centering
%     \begin{tabular}{l@{~}l}
%     \includegraphics[trim=0 1670 0 0, clip,width=0.48\linewidth]{results/women_segmentation.jpg}&    \includegraphics[trim=0 0 0 1670, clip,width=0.48\linewidth]{results/women_segmentation.jpg} \\
%   ~~~~~~ (a)   ~~~~~~~      (b) ~~~~~~~~ (c) ~~~~~~~~~ (d) ~~~~~~~~ (e) & ~~~~~~~ (a)   ~~~~~~~~~      (b) ~~~~~~~~ (c) ~~~~~~~ (d) ~~~~~~~ (e) 
%   \end{tabular}
%     \caption{Segmentation of women's hair. (a) original image, (b) ground truth segmentation, (c) our results, (d) the results of~\cite{ori}, (e) the results of~\cite{app8}}
%     \label{fig:segmentation_woman}
% \end{figure}

\subsection{Ablation Analysis}
\label{sec:ablation}

An ablation analysis is \ab{performed and reported quantitatively} in Tab.~\ref{tab:ablation_numerics}, and visually in appendix~\ref{sec:ablationapp}  for the task of facial hair content transfer. Without $\mathcal{L}^B_{Recon1}$ and $\mathcal{L}^A_{Recon1}$, the masks produced are empty and hence facial hair is not transferred to the target image, indicated by the high cosine similarity values but low classifier scores (i.e., the classifier labels the output as belonging to domain $A$). 
Similarly, without $\mathcal{L}_{DC}$, masks produced are empty as no disentanglement is possible.

Without   $\mathcal{L}_{Cycle}$ the masks produced include larger portions of the face, which also maintains similarity but hurts the classification score. $\mathcal{L}^B_{Recon2}$ and $\mathcal{L}^A_{Recon2}$ play a lesser role for the mask. Without $\mathcal{L}^B_{Recon2}$ the mask is less smooth, and without $\mathcal{L}^A_{Recon2}$ the mask still captures additional objects (e.g eyes). In fact, $\mathcal{L}^A_{Recon2}$ is a way to enforce the mask to capture the relevant content in a self-regularizing way. $\mathcal{L}^A_{Recon2}$ and $\mathcal{L}^B_{Recon2}$ are dependent on z, which is semantically aware of the domain specific content, while $L2$ equally penalizes any region of the image regardless of its content. When trying to use L2 norm, the mask had to be carefully adjusted to each experiment and resulted in a non-smooth mask which covers unnecessary parts of the face. This can be seen visually in appendix Fig~\ref{fig:ablation} and from the ``L2 reg'' entry of Tab.~\ref{tab:ablation_numerics}, where L2 regularization is used instead of $\mathcal{L}^A_{Recon2}$ and $\mathcal{L}^B_{Recon2}$. We note that \cite{atengan} uses sparsity regularization on the masks and \cite{atenguided} uses early stopping, which we do not require due to the regularization of $\mathcal{L}^A_{Recon2}$  and $\mathcal{L}^B_{Recon2}$. Refer to appendix~\ref{sec:additional_ablation} for further discussion.

%However, they are important for obtaining diversity in the generated images WHY AND ALSO SEE WHAT IS WITH THE FID.%A role of regularizing the mask is also played by $\mathcal{L}^B_{Recon2}$ and $\mathcal{L}^A_{Recon2}$, but to a lesser extent.  further reduce the size of the mask so as to cover only the facial hair. 

% \begin{table}[t]
% \caption{An ablation study showing the numerical effect of removing each of the losses}
%   \label{tab:ablation_numerics}
% \begin{center}
%   \begin{tabular}{lccc}
%     \toprule
%   	& Classifier & Similarity &  KID  \\ %& FID \\ %
%   	\midrule
%   	$\mathcal{L}$ & 88.1\% & $0.89$ & $3.5\pm0.1$  \\ % $90.7\pm1.8$ \\
%   	w/o $\mathcal{L}^A_{Recon2}$ & 88.5\% & $0.85$ &  $4.1\pm0.5$  \\ % $91.5\pm0.7$ \\
%   	w/o $\mathcal{L}^B_{Recon2}$ & 88.1\%& $0.87$ & $4.2\pm0.4$   \\ % $90.4\pm1.9$\\
%   	w/o $\mathcal{L}_{Cycle}$ &  67.1\% & $0.95$ & $4.1 \pm 0.9$ \\ %& $95.1 \pm 1.1$\\
%   	w/o $\mathcal{L}^B_{Recon1}$  & 9.4\% & $1.0$ & $4.3 \pm 0.7$ \\ %& $94.4 \pm 1.5$\\
%   	w/o $\mathcal{L}^A_{Recon1}$ & 9.7\% & $1.0$ & $4.6 \pm1.0$ \\ %& $96.9 \pm 3.9$\\
%     \bottomrule
% \end{tabular}
% \end{center}
% \end{table}

% \begin{figure}
% \centering
% \begin{tabular}{cc}
%   \includegraphics[width=0.5\linewidth,clip]{results/lfw_2lines.jpg} &
%   \includegraphics[width=0.5\linewidth,clip]{results/alien_2lines.jpg} \\
%   (a) & (b)\\
%   \end{tabular}
% \caption{Out of domain images. (a) Results obtained by manipulating LFW images. (b) Results on extremely out of domain images. }
%   \label{fig:out}
% \end{figure}

\section {Conclusions}

When transferring content between two images, we need to know what to transfer, where to transfer it to, and how to transfer it. Previous work in guided transfer either transferred global style properties or neglected the ``where'' aspect, which ultimately lead to an ineffective generation that lacks attention.

In our work, the ``what'' aspect is captured by $E_s$, and $D_B$ captures both the ``where'' and the ``how''. Our results demonstrate that the context (image $a$) in which the content is placed determines not just the location of the inserted content but also the form in which it is presented, where both aspects can vary dramatically, even for a fixed content-guide image $b$. The comprehensive modeling of the guided content transfer problem leads to results that are far superior to the current state of the art. In addition, the modelling of ``where'' allows us to obtain accurate segmentation masks in a weakly supervised way, remove content, swap content between images, and add multiple contents without suffering a gradual degradation in quality.

\section*{Acknowledgements}
This project has received funding from the European Research Council (ERC) under the European
Union’s Horizon 2020 research and innovation programme (grant ERC CoG 725974).

%\newpage

%\bibliographystyle{neurips_2019}
\bibliographystyle{iclr2020_conference}
\bibliography{gans}

\appendix 

\section{Architecture and Hyperparameters}\label{sec:arc}

We consider samples in $A$ and $B$ to be images in $\mathbb{R}^{3\times 128\times 128}$.
The encoders $E_c$ and $E_s$ each consist of $6$ convolutional blocks. Similarly, $D_A$ and $D_B$ consist of $6$ de-convolutional blocks. 

A convolutional block $d_k$ consisting of: (a) $4\times4$ convolutional layer with stride $2$, pad $1$ and $k$ filters  (b) a batch normalization layer (c) a Leaky ReLU activation with slope $0.2$. Similarly, a de-convolutional block $u_k$ consists of: (a) $4\times4$ de-convolutional layer with stride $2$, pad $1$ and $k$ filters   (b) a batch normalization layer (c) a ReLU activation.

The structure of the encoders and decoders is then: $E_c\text{: }  d_{32}, d_{64}, d_{128}, d_{256}$, $d_{512-sep}, d_{512-2\cdot sep}$, $E_s\text{: } d_{32}, d_{64}, d_{128}, d_{128}$, $d_{128}, d_{sep}$, $D_A\text{: } u_{512}, u_{256}$, $u_{128}, u_{64}$, $u_{32}, u^*_{3}$, and $D_B\text{: } u_{512}$, $u_{256}, u_{128}$, $u_{64}, u_{32}, u^*_{4}$.

The last layer of $D_A$ ($u^*_{3}$) differs in that it does not contain batch normalization and tanh activation is applied, instead of ReLU. $D_B$'s last layer ($u^*_{4}$) similarly does not contain batch normalization. The output is of size $4\times 128\times 128$. We split the output to a mask (first channel) and raw output (other three channels). We apply a sigmoid activation to the mask to get values between $0$ and $1$ and a tanh activation for the raw output and a Tanh activation for the raw output. $sep$ is the dimension of the separate encoders, set to be $100$ for all datasets.

The discriminator $C$ consists of a fully connected layer of $512$ filters, a Leaky ReLU activation with slope $0.2$, a second fully connected layer of one filter and a final sigmoid activation.

We use the Adam optimizer  with $\beta_1 = 0.5, \beta_2 = 0.999$, and 
learning rate of $0.0002$. We use a batch size of size $32$ in training. 

We constructed the train/test sets using 90\%-95\% split.
This consists of about 7,200-18,000 examples for train and about 800-2,000 examples for test for each attribute.

\section{Additional Content Transfer Results}\label{sec:add_res}

Additional results to the ones presented in the main text are provided here. 

\begin{figure*}[t]
\centering
\begin{tabular}{c}
  \includegraphics[width=0.95\linewidth, clip]{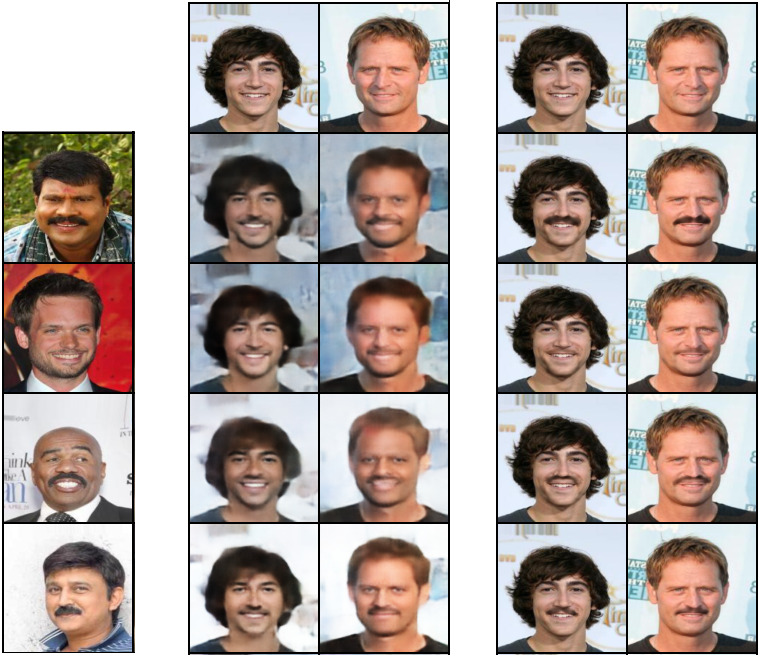} \\
  \hspace{-1.2cm} (a)   ~~~~~~~~~~~~~~~~~~~~~~~~~~~~~~~~~~~~~~~~~~      (b) ~~~~~~~~~~~~~~~~~~~~~~~~~~~~~~~~~~~~~~~~~~~~~~~~~~~~~~~~ (c)
  \end{tabular}
\caption{(a) Guide images in domain $B$ (faces with facial hair). (b) Results by the method of \cite{ori}: the  top row is the source images in domain $A$. The others incorporate the facial hair from the corresponding row of (a). (c) Same mapping for our method.}
  \label{fig:mustach_comapre}
\end{figure*}

\begin{figure*}[t]
    \centering
    \includegraphics[width=0.95\linewidth]{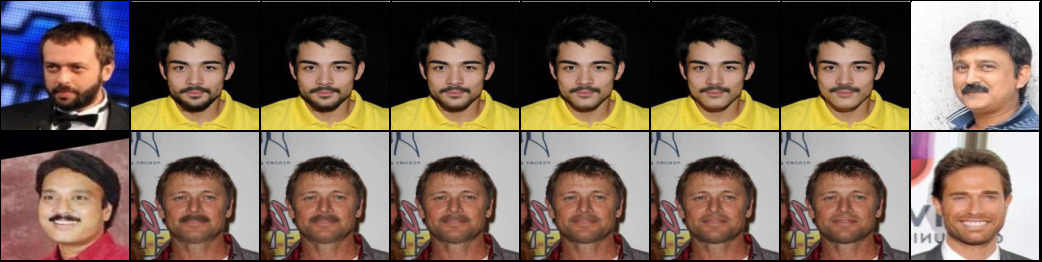} 
    \caption{Facial Hair Interpolation}
    \label{fig:mustache_in}
\end{figure*}

\begin{figure*}[t]
    \centering
    \includegraphics[width=0.95\linewidth]{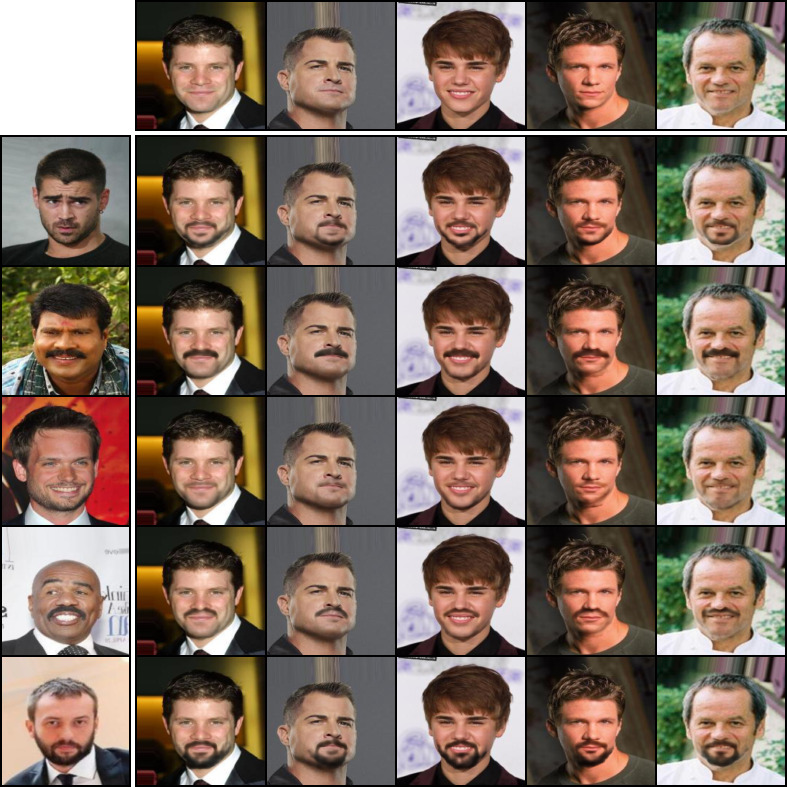} 
    \caption{Additional results for the guided transfer of facial hair. }
    \label{fig:mustach_transfer}
\end{figure*}

\begin{figure*}[t]
    \centering
    \includegraphics[width=0.95\linewidth]{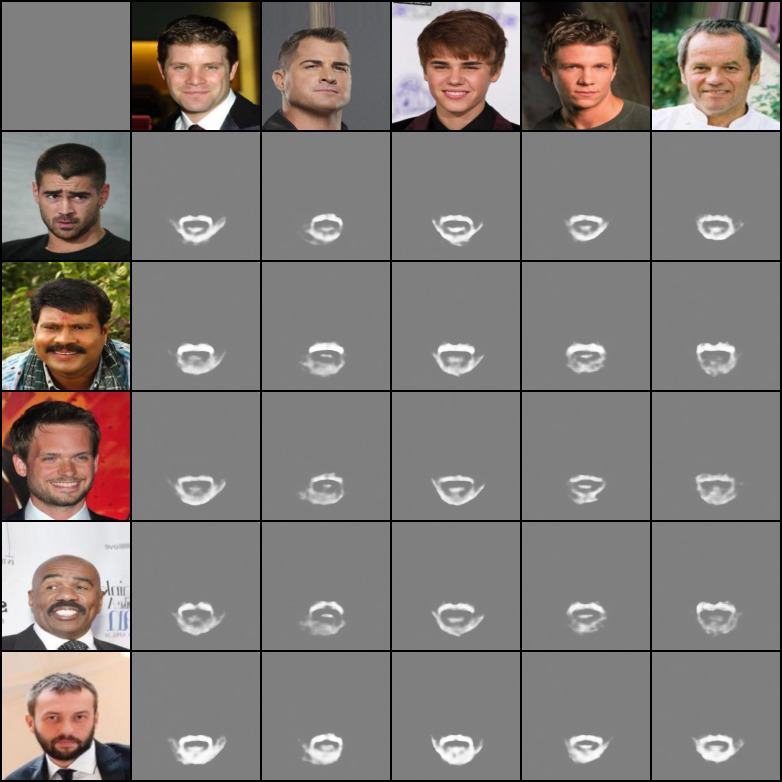} 
    \caption{Masks generated for the guided transfer of facial hair experiment. Masks generated are for the translated images in Fig~\ref{fig:mustach_transfer}.}
    \label{fig:mustach_masks}
\end{figure*}

\begin{figure*}[t]
    \centering
    \includegraphics[width=0.95\linewidth]{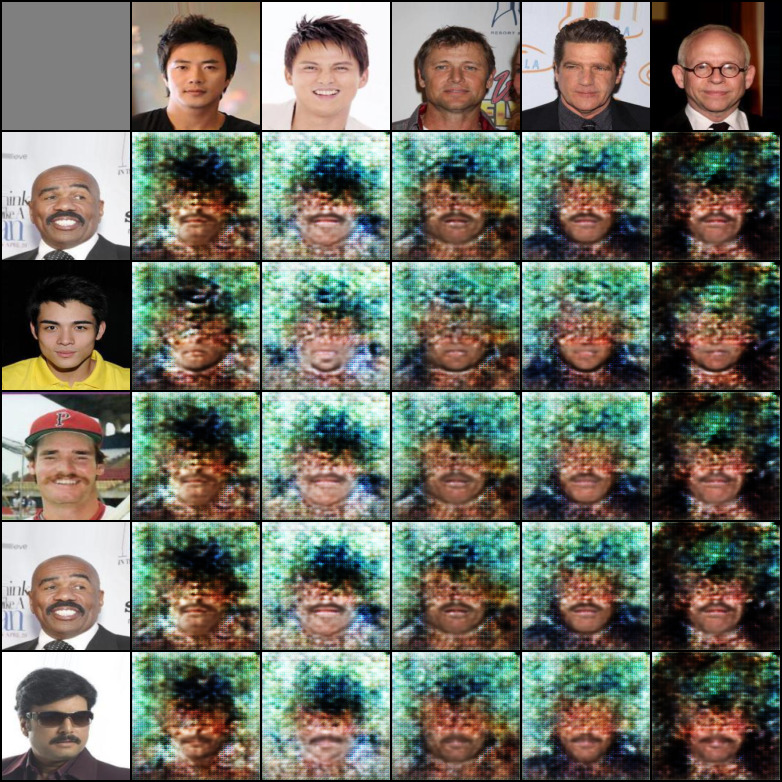} 
    \caption{Raw outputs generated by $D_b$ for the task of facial hair content transfer.}
    \label{fig:row_o}
\end{figure*}

Fig.~\ref{fig:mustach_comapre} gives a comparison of our method to the state-of-the-art for the transfer of facial hair.
%, in addition to Fig.~3 of the main text.
Fig.~\ref{fig:mustache_in} provides additional interpolation results for this task while Fig.~\ref{fig:mustach_transfer} provides additional content transfer results. Fig.~\ref{fig:mustach_masks} shows the masks generated for this content transfer. Fig.~\ref{fig:row_o} gives an example of the raw output given by our method for this task. 

\begin{figure*}[t]
    \centering
    \includegraphics[width=0.95\linewidth]{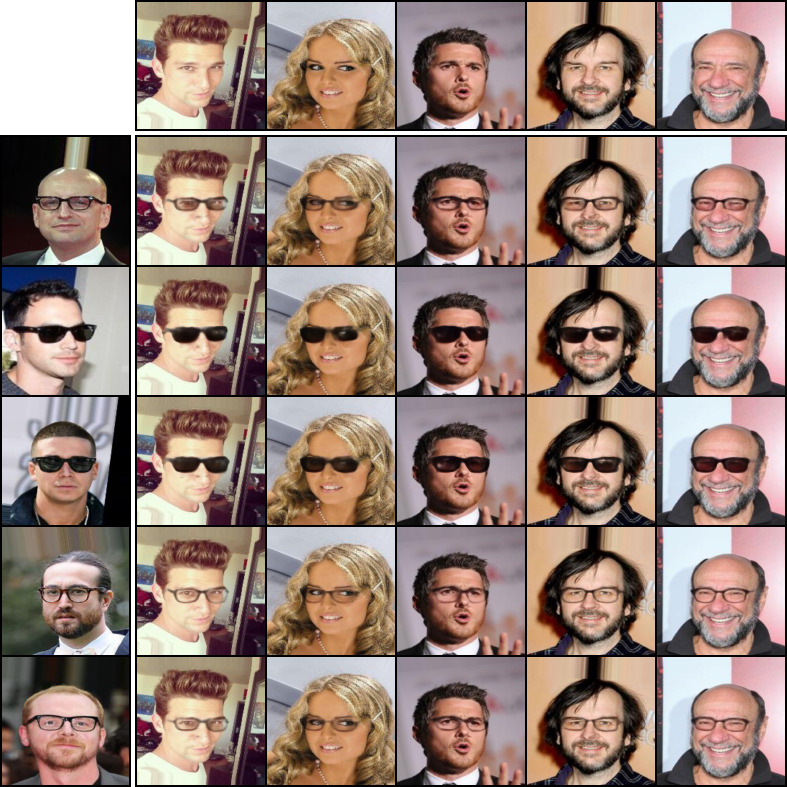} 
    \caption{Additional content transfer example. Given an image with glasses (left), and another image of a face with no glasses (top), the proposed method identifies and translates the specified glasses from the former domain to the latter. }
    \label{fig:glasses_transfer}
\end{figure*}

\begin{figure*}[t]
    \centering
    \includegraphics[width=0.95\linewidth]{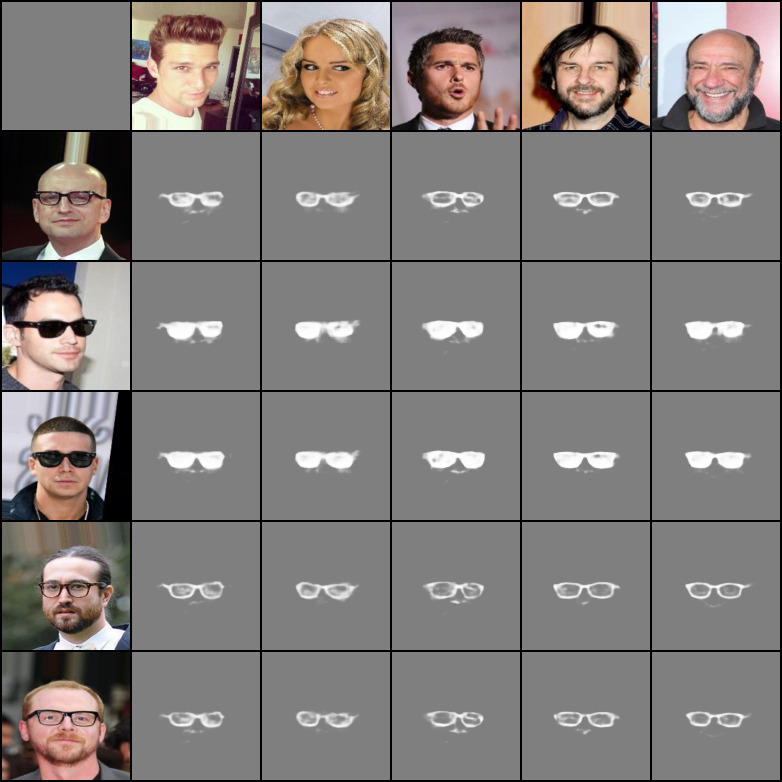} 
    \caption{Masks generated for the guided transfer of glasses experiment. Masks generated are for the translated images in Fig~\ref{fig:glasses_transfer}.}
    \label{fig:glasses_masks}
\end{figure*}

\begin{figure*}[t]
\centering
\begin{tabular}{c}
  \includegraphics[width=0.95\linewidth, clip]{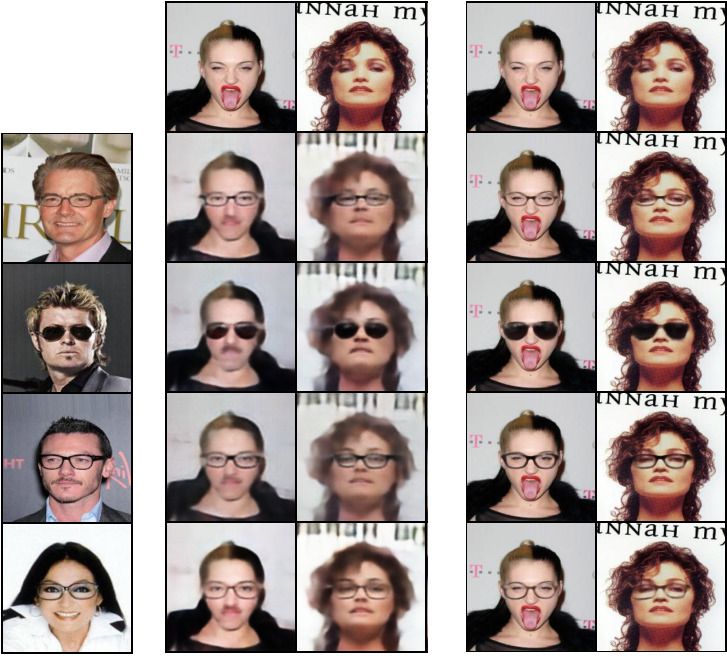} \\
  \hspace{-1.2cm} (a)   ~~~~~~~~~~~~~~~~~~~~~~~~~~~~~~~~~~~~~~~~~~      (b) ~~~~~~~~~~~~~~~~~~~~~~~~~~~~~~~~~~~~~~~~~~~~~~~~~~~~~~~~ (c)
  \end{tabular}
\caption{(a) Guide images in domain $B$ (glasses). (b) \cite{ori} method: the top row is the source images in domain $A$. The others incorporate the glasses from the corresponding row of (a). (c) Same mapping for our method.}
  \label{fig:glass_comapare}
\end{figure*}

Fig.~\ref{fig:glasses_transfer} gives additional results for the task of glasses transfer, while Fig.~\ref{fig:glasses_masks} shows the masks generated for this content transfer. Fig.~\ref{fig:glass_comapare} provides additional comparison to the baseline method. 

\begin{figure*}[t]
    \centering
    \includegraphics[width=0.95\linewidth]{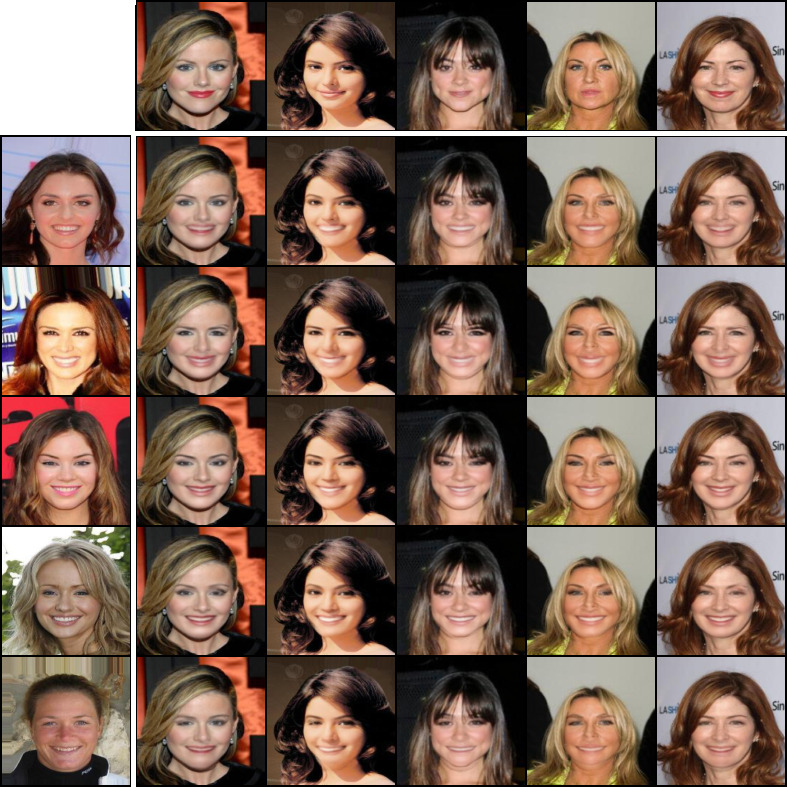} 
    \caption{Additional content transfer example. Given an image of smiling face (left), and another image of a non-smiling face (top), the proposed method identifies and translates the specified smiles from the former domain to the latter. }
    \label{fig:mouth_ex}
\end{figure*}

\begin{figure*}[t]
    \centering
    \includegraphics[width=0.95\linewidth]{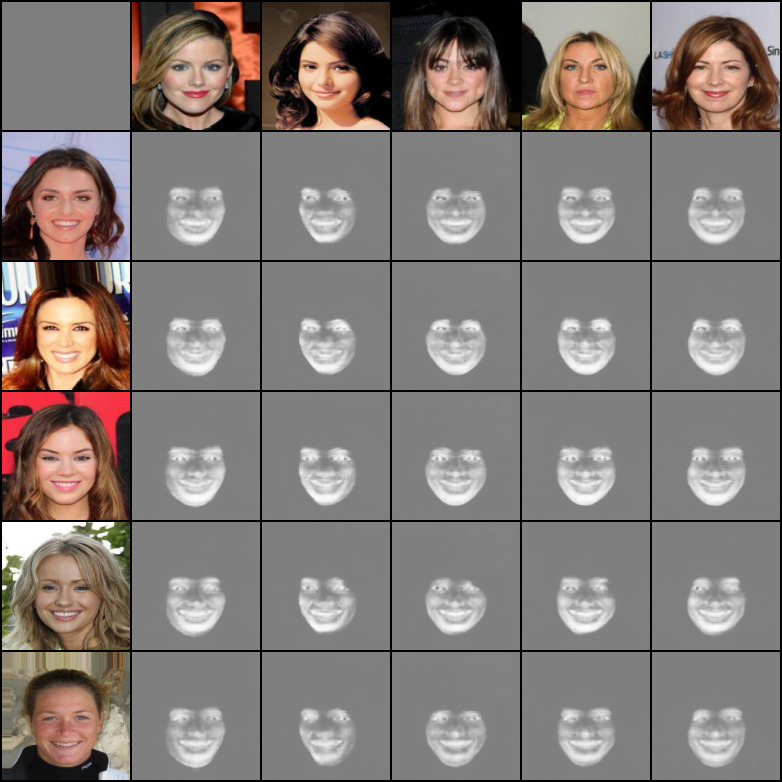} 
    \caption{Masks generated for the guided transfer of smile experiment. Masks generated are for the translated images in Fig~\ref{fig:mouth_ex}. }
    \label{fig:mouth_mask}
\end{figure*}

\begin{figure*}[t]
    \centering
    \includegraphics[width=0.95\linewidth]{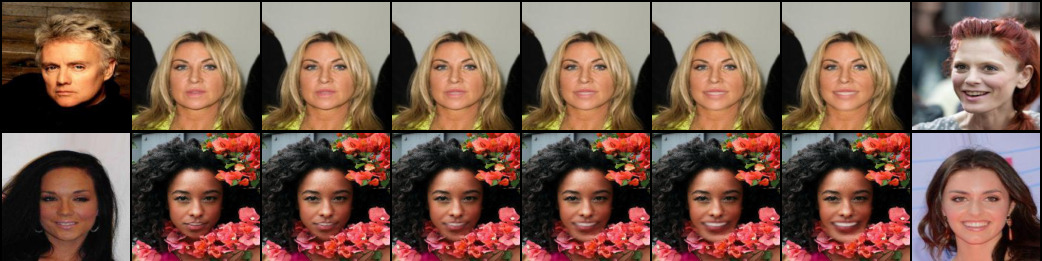} 
    \caption{Smile Interpolation}
    \label{fig:mouth_in}
\end{figure*}

\begin{figure*}[t]
\centering
\begin{tabular}{c}
  \includegraphics[width=0.95\linewidth, clip]{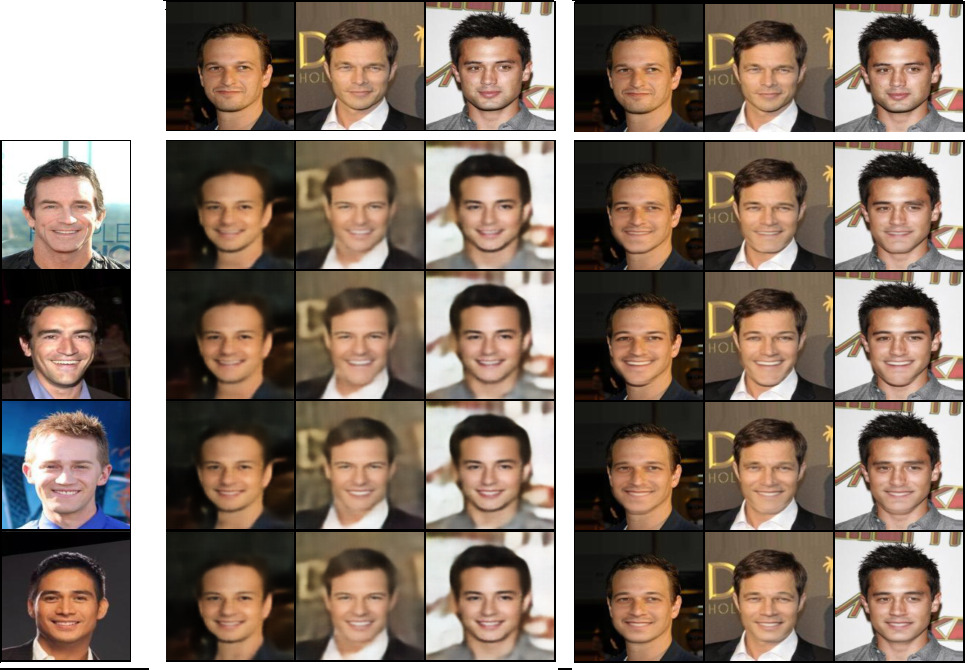} \\
  \hspace{-1.2cm} (a)   ~~~~~~~~~~~~~~~~~~~~~~~~~~~~~~~~~~~~~~~~~~      (b) ~~~~~~~~~~~~~~~~~~~~~~~~~~~~~~~~~~~~~~~~~~~~~~~~~~~~~~~~ (c)
  \end{tabular}
\caption{(a) Guide images in domain $B$ (faces with smile). (b) \cite{ori}: the  top row is the source images in domain $A$. The others incorporate the smile from the corresponding row of (a). (c) Same mapping for our method.
  \label{fig:smile_comapare}}
\end{figure*}

Fig.~\ref{fig:mouth_ex} and Fig.~\ref{fig:mouth_mask} provide additional content transfer results and the generated masks for the task of smile transfer. 
It is well known that smile includes not only the mouth but also other facial features such as eyebrows and cheeks \cite{smile_citation}, thus when our method transfer the smile, it transfer all the relevant facial features for the smile as can be seen in the generated masks in Fig.~\ref{fig:mouth_mask}. Fig.~\ref{fig:mouth_in} provide interpolation results and Fig.~\ref{fig:smile_comapare} gives a comparison of our method for this task.

\begin{figure*}[t]
    \centering
    \includegraphics[width=0.95\linewidth]{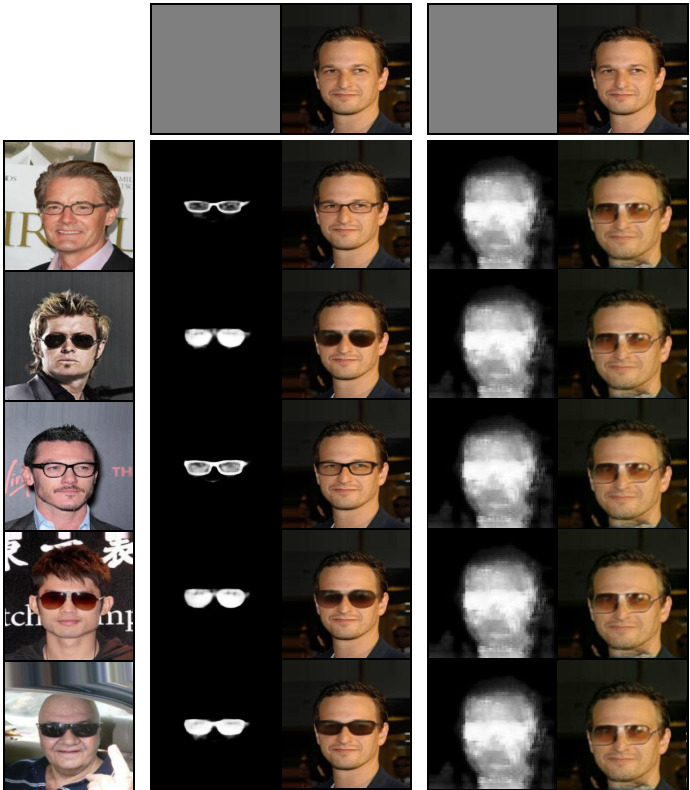} 
    \caption{Our method (left) compared to \cite{atenguided} (right) on the task of adding glasses to the original image (top). We show the generated masks and the final result for both methods. For our method we also show the guidance images. }
    \label{fig:compare_25}
\end{figure*}

Fig~\ref{fig:compare_25} gives a comparison of our method to \cite{atenguided}. While our method uses guidance image which allows one to many translation, \cite{atenguided} can translate to only one image.

\section{Additional out of domain manipulations results}
\label{sec:transfer}

\begin{figure}
\centering
\begin{tabular}{c}
  \includegraphics[width=0.95\linewidth,clip]{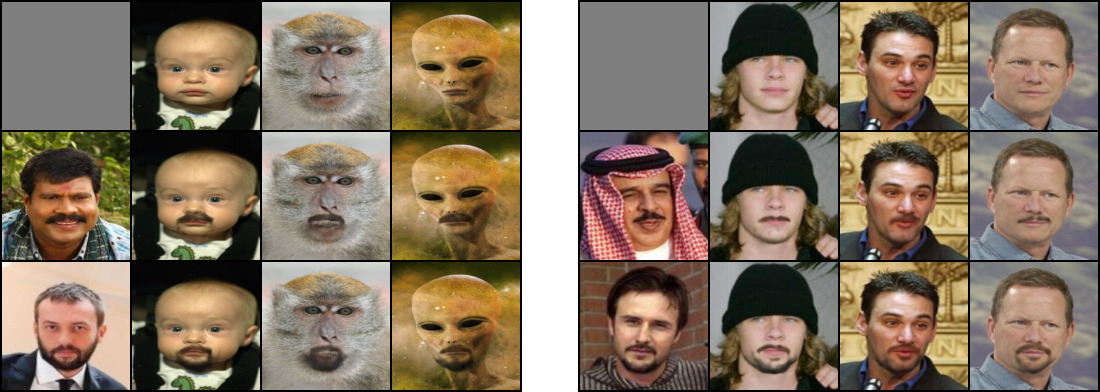} \\
  \hspace{-1.2cm} (a)  ~~~~~~~~~~~~~~~~~~  ~~~~~~~~~~~~~~~~~~~~~~~~~~~~~~~~~~~~~~~~~~      (b)
  \end{tabular}
\caption{Out of domain translation. (a) Results on extremely out of domain images. (b) Results obtained by manipulating LFW images.  }
  \label{fig:transfer}
\end{figure}

\begin{figure*}[b]
\centering
\begin{tabular}{c}
  \includegraphics[width=0.95\linewidth, clip]{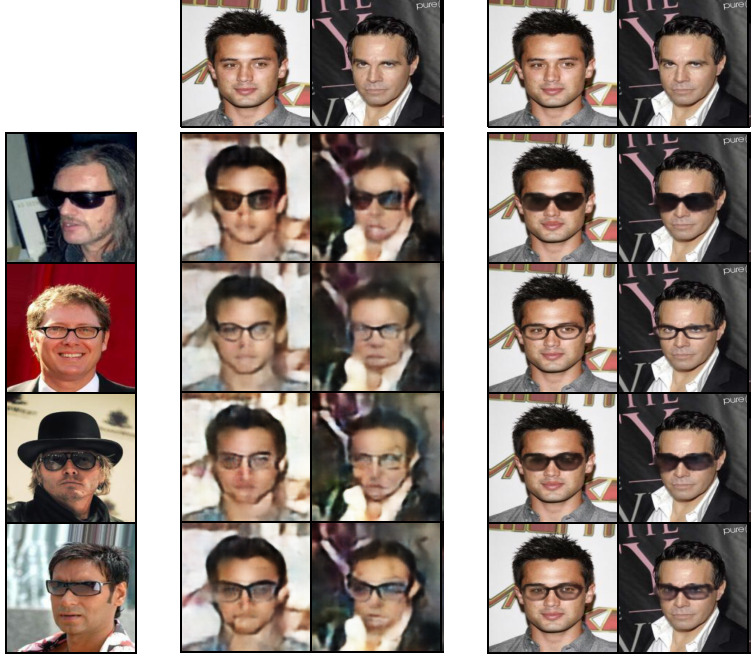} \\
  \hspace{-1.2cm} (a)   ~~~~~~~~~~~~~~~~~~~~~~~~~~~~~~~~~~~~~~~~~~      (b) ~~~~~~~~~~~~~~~~~~~~~~~~~~~~~~~~~~~~~~~~~~~~~~~~~~~~~~~~ (c)
  \end{tabular}
\caption{Out of distribution translation. The mapping between faces without and with glasses is trained only on women and applied to men. (a) Guide images in domain $B'$ (men with glasses) and top row is the source images in domain $A'$ (men without glasses), both not given during training. (b) The remaining rows are translated images by the method of \cite{ori}. (c) Same mapping for our results.}
  \label{fig:transfer_glasses_compaer}
\end{figure*}

\begin{figure*}[b]
\centering
\begin{tabular}{c}
  \includegraphics[width=0.95\linewidth, clip]{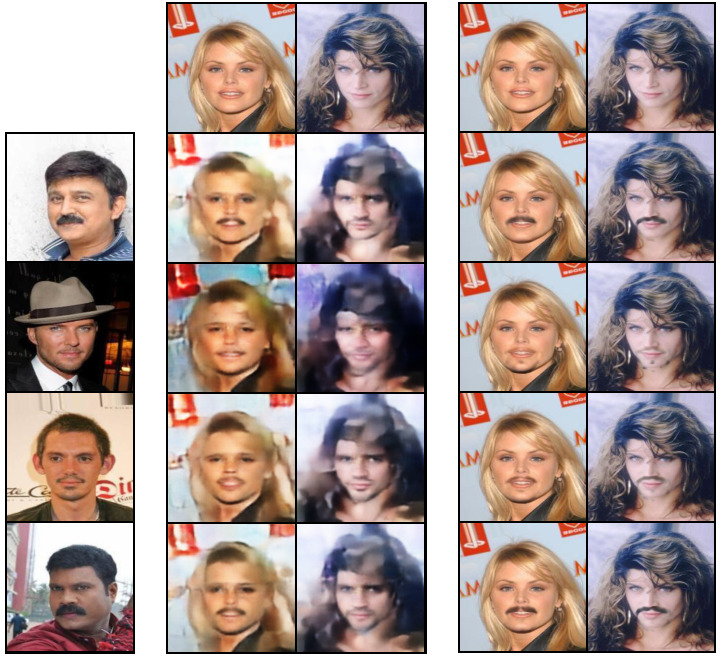} \\
  \end{tabular}
\caption{Additional out of distribution translation results. We train on mapping facial hair from male faces (left) to male faces, and apply this to women's faces (top) during inference time. Our domain translation results (right) are compared to those of \cite{ori} (middle).}
  \label{fig:mustach_transfer_women}
\end{figure*}

Fig.~\ref{fig:mustach_transfer_women} shows sample results where the mapping of facial hair is applied to female faces. Out of distribution translation where the train domains are different from the inference domains are shown for glasses in Fig~\ref{fig:transfer_glasses_compaer}. We further consider the ability of our method to perform translation on images from the out-of-distribution LFW dataset \cite{lfw} (Fig.\ref{fig:transfer}(a)), as well as images of an alien and a baby (Fig.\ref{fig:transfer}(b)) which are extremely out of distribution, all not present during training. As can be seen, even in these cases, our method successfully transfers the desired content.
.

\section{Comparative results for the handbag dataset}
\label{sec:handbags}
\begin{figure*}[t]
\centering
\begin{tabular}{c}
  \includegraphics[width=0.85\linewidth, clip]{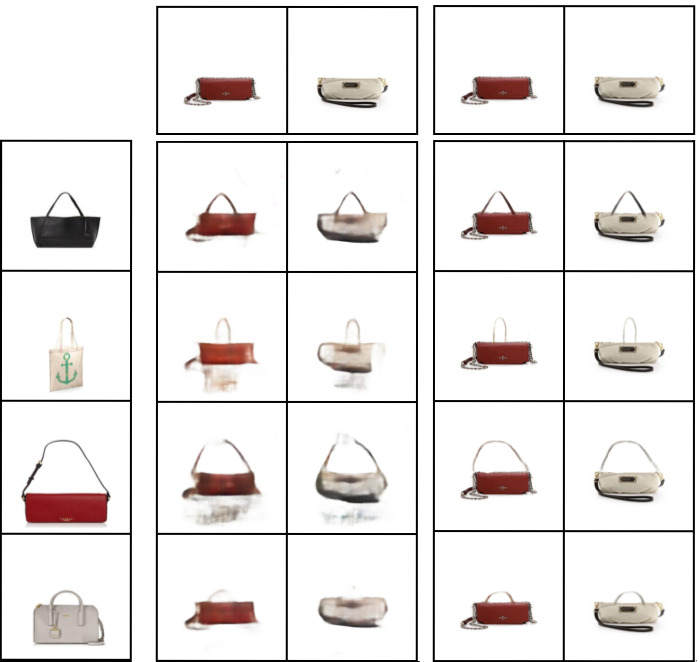} \\
  \hspace{-1.2cm}~~~~~ (a)   ~~~~~~~~~~~~~~~~~~~~~~~~~~~~~~~~~~~~~~      (b) ~~~~~~~~~~~~~~~~~~~~~~~~~~~~~~~~~~~~~~~~~~~~~~~~ (c)
  \end{tabular}
\caption{(a) Guide images in domain $B$ (handbags with handles). (b) \cite{ori}: the  top row is the source images in domain $A$. The others incorporate the handles from the corresponding row of (a). (c) Same mapping for our method.}
  \label{fig:bag_comapare}
\end{figure*}

\begin{figure*}[t]
    \centering
    \includegraphics[width=0.85\linewidth]{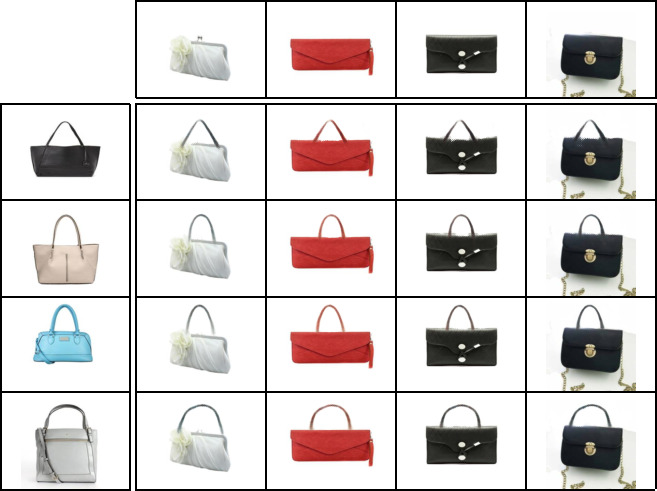} 
    \caption{Additional content transfer example. Given an image of bag with a handle (left), and another image of a handbag (top), the proposed method identifies and translates the specified handbag from the former domain to the latter. }
    \label{fig:hanbag_grid}
\end{figure*}

Fig.~\ref{fig:bag_comapare} gives a comparison of our method for the handle transfer for handbags, while Fig.~\ref{fig:hanbag_grid} provides more  results for this task (on top of Fig.~\ref{fig:handbags}). 

\section{A qualitative comparison of our method to literature methods on the attribute removal task}
\label{sec:removal}

\begin{figure*}[t]
\centering
    \includegraphics[width=0.92\linewidth, clip]{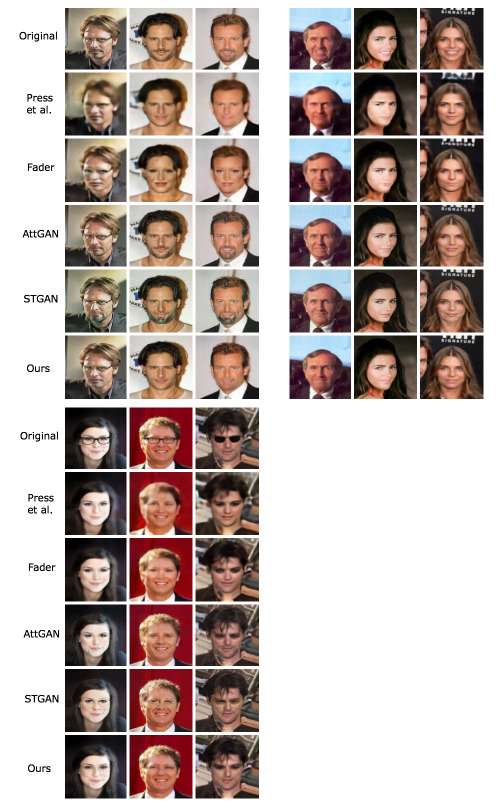}
\caption{Attribute removal for the task of mustache (top left), smile (top right) and glasses (bottom left). The result of our method is shown alongside the baseline methods, \cite{ori}, Fader (\cite{lample2017fader}), AttGAN (\cite{attgan}) and STGAN (\cite{stgan}). }
  \label{fig:removal}
\end{figure*}

Fig.~\ref{fig:removal} gives a comparison of our method on the task of attribute removal. The translation of \cite{ori} is blurry and suffers from many of the facial features being lost. For example, for glasses removal, the men on the right have facial hair which is lost in the translation.  \cite{lample2017fader} completely changes the facial features. For example, for facial hair removal, the gender seems to change from men to women. \cite{attgan} and \cite{stgan} are unable to remove the mustache and the translation is of lower quality in general. For example, for the glasses removal, for the man in the middle, the translation is unnatural around the eyes.  Our translation is of consistently higher quality for all tasks and successfully removes the desired attribute.

\section{Additional sequential content transfer and attribute removal results} \label{sec:two_att}

We provide additional images produced by our method as well as by the baseline method. As can be seen, in order to perform the guided content transfer of two attributes from two different domain, \cite{ori} passes the source input image into the network twice which wastefully reconstructs static facial features twice. 

\begin{figure*}[t]
\centering
\begin{tabular}{c}
  \includegraphics[width=0.95\linewidth, clip]{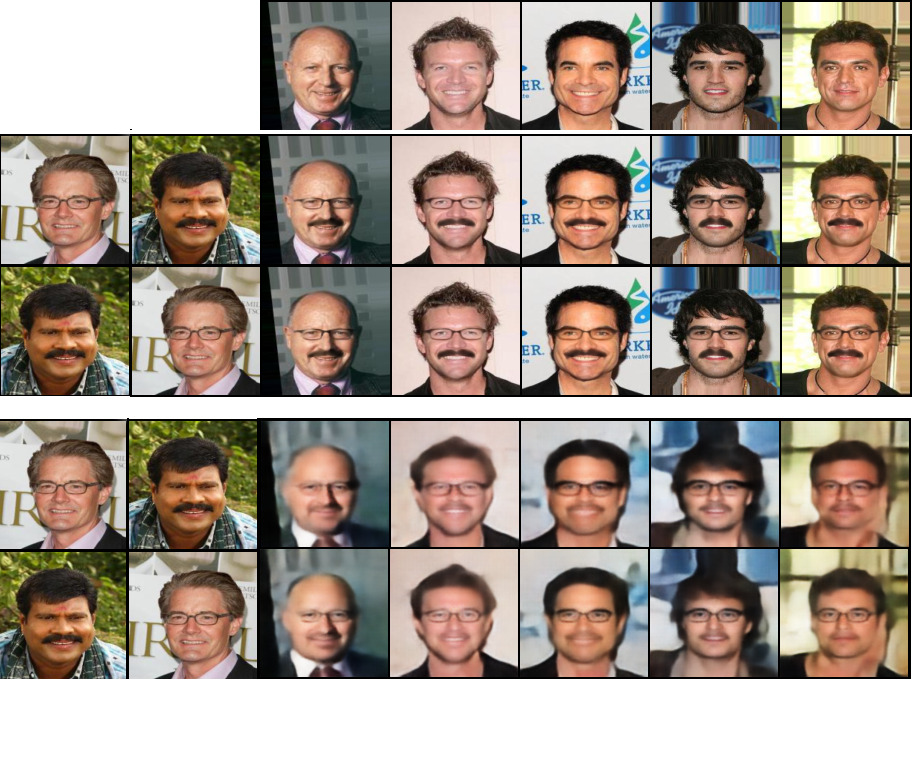} 
  \end{tabular}
\caption{First two images on the left are the content donors applied sequentially, either with facial hair or glasses. The top row is the input source images. The results on the bottom are the translation of \cite{ori} while on the top are our results.}
  \label{fig:two_att_transfer}
\end{figure*}

For example, for sequential addition of glasses and facial hair, as seen in Fig~\ref{fig:two_att_transfer}, our method successfully transfers the two attributes, while  \cite{ori} not only produces blurry images, but is much worse at transferring the content from both attributes. This observation is also supported by the user study performed in Tab~\ref{tab:user_study}.

\begin{figure*}[t]
    \centering
    \includegraphics[width=0.95\linewidth]{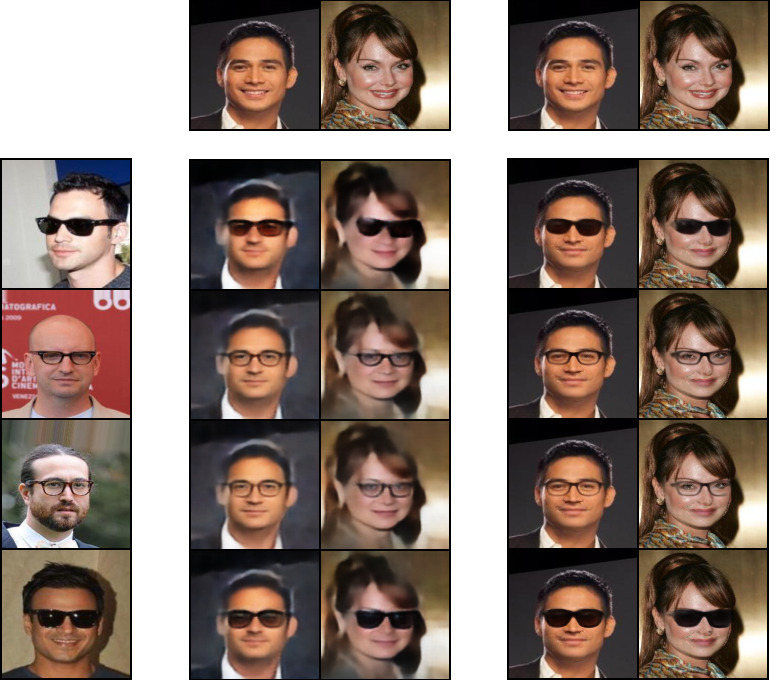} 
    \caption{Additional removal and content transfer results. Given an image with glasses (left), and another image of a face with no glasses and a smile (top), the proposed method removes the smile and identifies and translates the specified glasses from the former domain to the latter. In the middle are the translated examples of \cite{ori} while on the right are our translated results.}
    \label{fig:add_rm_compare}
\end{figure*}

\begin{figure*}[t]
    \centering
    \includegraphics[width=0.95\linewidth]{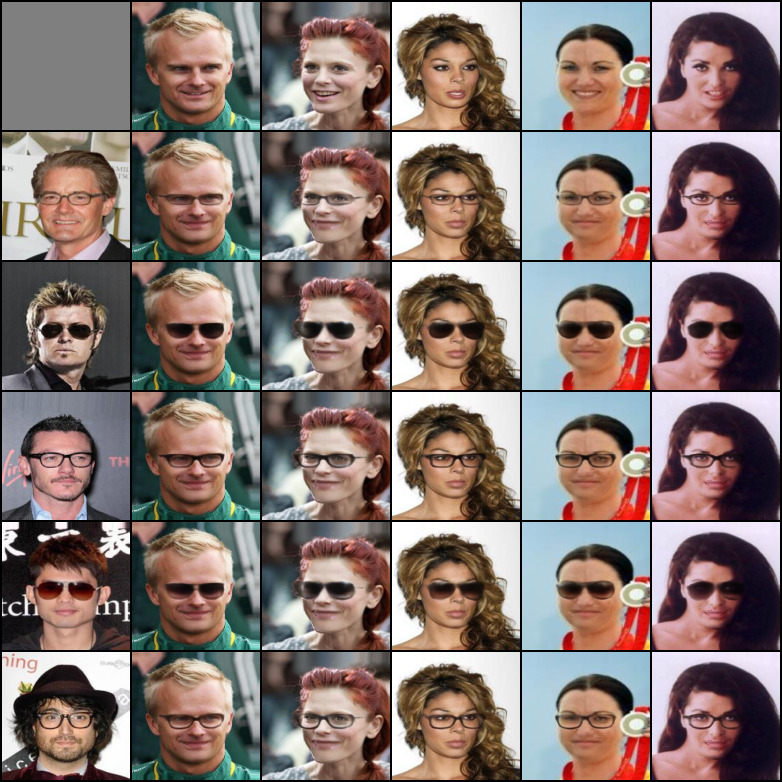} 
    \caption{{Additional removal and content transfer results for smile removal and glasses addition. Given an image with glasses (left), and another image of a face with no glasses and a smile (top), the proposed method removes the smile and translates the specified glasses from the former domain to the latter.}}
    \label{fig:add_rm}
\end{figure*}

\begin{figure*}[t]
    \centering
    \includegraphics[width=0.95\linewidth]{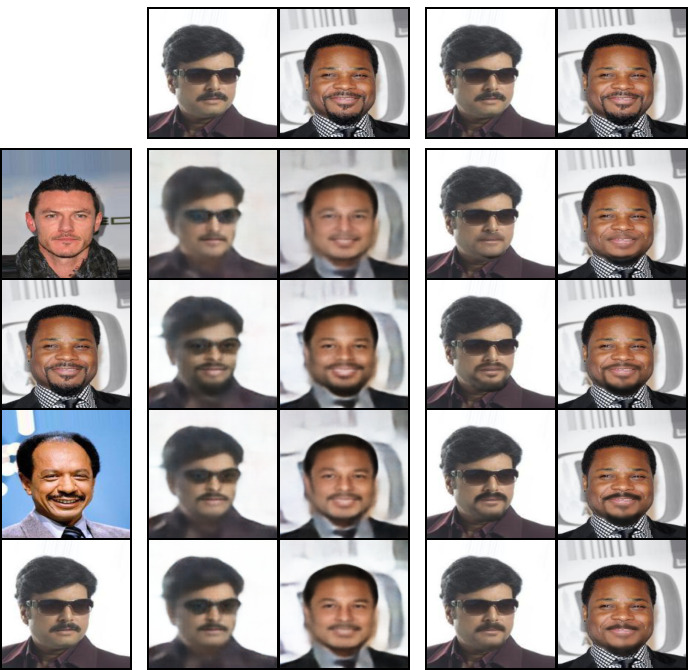} 
    \caption{\textcolor{black}{Facial hair swap results. Our method first removes the facial hair and then adds the facial hair of the guided image on the left. In the middle are the translated examples of \cite{ori} while on the right are our translated results.}}
    \label{fig:replace_mustache_compare}
\end{figure*}

\begin{figure*}[t]
    \centering
    \includegraphics[width=0.95\linewidth]{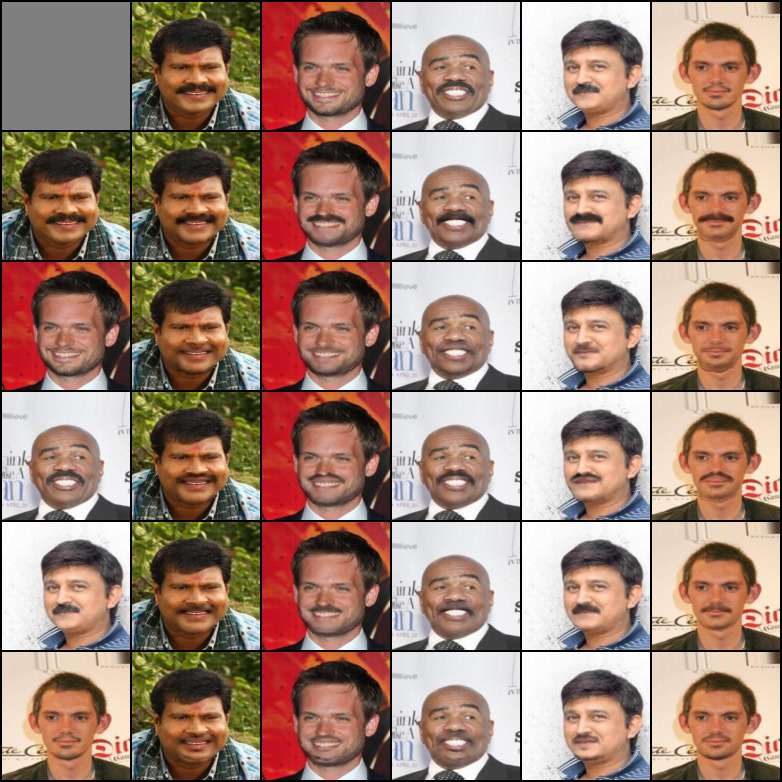} 
    \caption{Additional facial hair swap results. Our method first removes the facial hair and then adds the facial hair of the guided image on the left}
    \label{fig:replace_mustache}
\end{figure*}

Fig.~\ref{fig:add_rm_compare} show the comparison to the baseline model for the task of closing the mouth and adding glasses; Fig.~\ref{fig:add_rm} shows additional results from our method for this task. Fig.~\ref{fig:replace_mustache_compare} provides the comparison for the task of replacing facial hair; while Fig.~\ref{fig:replace_mustache} presents additional results for this task.

\section{Additional weakly supervised segmentation results}
\label{sec:segment}

\begin{figure*}[t]
    \centering
    \begin{tabular}{c}
    \includegraphics[width=0.95\linewidth]{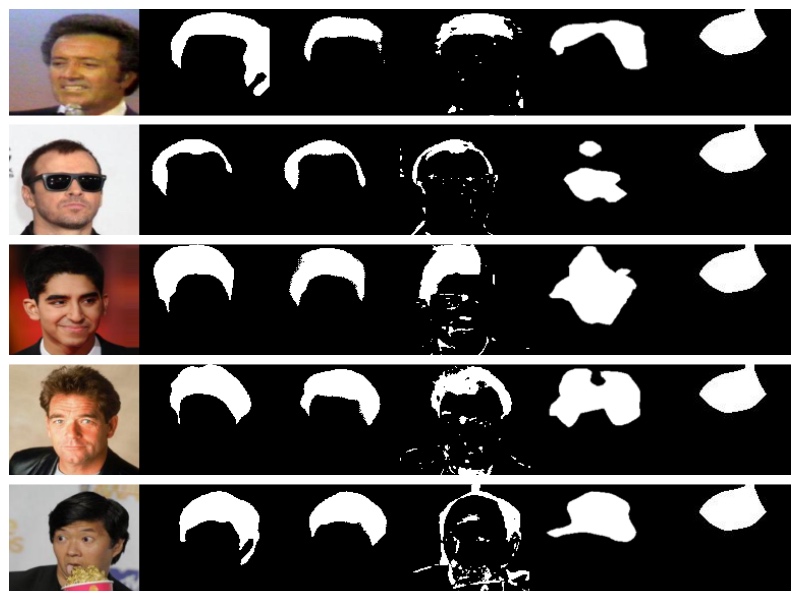} \\
  \hspace{-1.2cm} ~~~~~~~~~~~~~~~~~~~~~~~~ (a)   ~~~~~~~~~~~~~~~~~~      (b) ~~~~~~~~~~~~~~~~~~ (c) ~~~~~~~~~~~~~~~~~~ (d) ~~~~~~~~~~~~~~~~~~ (e)~~~~~~~~~~~~~~~~~~ (f)~~~~~~~~~~~
  \end{tabular}
    \caption{Segmentation of men's hair. (a) original image, (b) ground truth segmentation, (c) our results, (d) the results of~\cite{ori}, (e) the results of~\cite{app8}, (f) results of CAM.}
    \label{fig:seg_hait2}
\end{figure*}

\begin{figure*}[t]
    \centering
    \begin{tabular}{c}
        \includegraphics[width=0.95\linewidth]{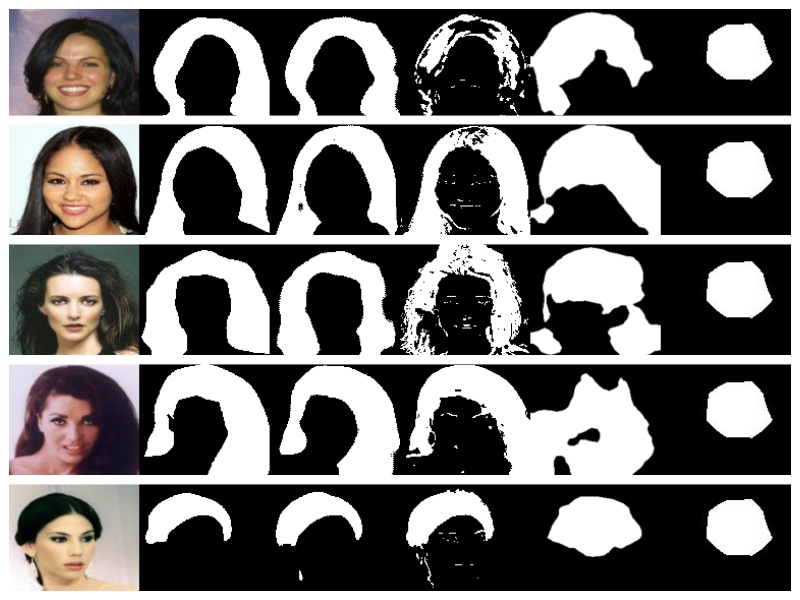} \\
  \hspace{-1.2cm} ~~~~~~~~~~~~~~~~~~~~~~~~ (a)   ~~~~~~~~~~~~~~~~~      (b) ~~~~~~~~~~~~~~~~~~ (c) ~~~~~~~~~~~~~~~~~~ (d) ~~~~~~~~~~~~~~~~~~ (e)~~~~~~~~~~~~~~~~~~ (f)~~~~~~~~~~~
  \end{tabular}
    \caption{Additional Segmentation results for of women's hair. (a) original image, (b) ground truth segmentation, (c) our results, (d) the results of~\cite{ori}, (e) the results of~\cite{app8}, (f) results of CAM.}
    \label{fig:seg_hait}
\end{figure*}

Fig~\ref{fig:seg_hait2} gives a comparison our method for the task of men's hair segmentation as given in section 4.2 of the main text, while Fig~\ref{fig:seg_hait} gives additional results for the segmentation of woman's hair.

\begin{figure*}[t]
    \centering
    \includegraphics[width=0.95\linewidth]{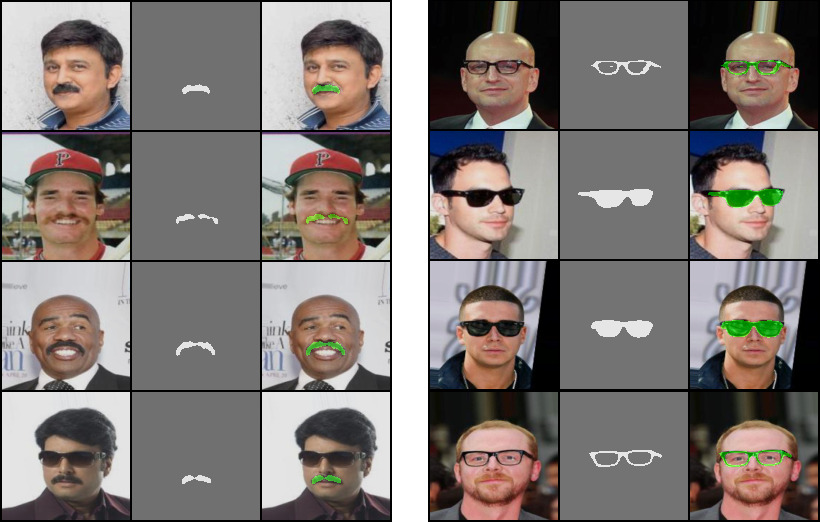}
    \caption{Additional segmentation results for the domain of glasses and facial hair. }
    \label{fig:seg2}
\end{figure*}

Additional segmentation results are shown in Fig.~\ref{fig:seg2} for the domain of glasses and facial hair. In this domain quantitative results cannot be obtained due to lack of ground truth segmentations.

\section{additional ablation study discussion}
\label{sec:additional_ablation}

While the loss in Eq.~\ref{eq:7} directly affects the mask generation, the losses in Eq.~\ref{eq:3} ($\mathcal{L}_{DC}$) and Eq.~\ref{eq:5} ($\mathcal{L}_{Recon1}^A$) affect it indirectly. Without the loss in Eq.~\ref{eq:3} ($\mathcal{L}_{DC}$), no disentanglement is possible, and the common encoder would contain all of the image information including the separate information. This means that the image produced by $D_A(E_c(b))$ is close to $b$ and, therefore, the generated mask is empty. 

Furthermore, without the loss of Eq.~\ref{eq:5} ($\mathcal{L}_{Recon1}^A$), we empirically observe that $D_A(E_c(b))$ outputs the image with the specific part intact (for example, the facial hair is not removed). This indirect effect on the disentanglement probably stems from the fact that without this loss, there is reconstruction only on faces with facial hair (running example for the specific part). Thus, $E_c$ can encode generic facial hair information for shaved faces and have $E_c(b)$ and $E_c(a)$ still indistinguishable. Eq.~\ref{eq:5} ($\mathcal{L}_{Recon1}^A$) makes sure that $E_c$ won’t encode facial hair for shaved faces, since it requires reconstruction of an image without facial hair. 

We further consider the effect of replacing the norm used for reconstruction losses from L1 to L2. When using L2 norm for $\mathcal{L}^B_{Recon1}$ and $\mathcal{L}^A_{Recon1}$ the result is comparable both numerically (``L2 recon \#1" in Tab.~\ref{tab:ablation_numerics}) and visually (see appendix Fig~\ref{fig:ablation}). When using L2 norm for $\mathcal{L}^B_{Recon2}$ and $\mathcal{L}^A_{Recon2}$  (``L2 recon \#2" in Tab.~\ref{tab:ablation_numerics}) the results change more significantly: the size of the mask is larger and the classifier score significantly lower. We attribute this to the sparsity inducing effect of the L1 

\section{Visual results of the ablation study}
\label{sec:ablationapp}
\begin{figure*}[t]
    \centering
    \includegraphics[width=0.75\linewidth]{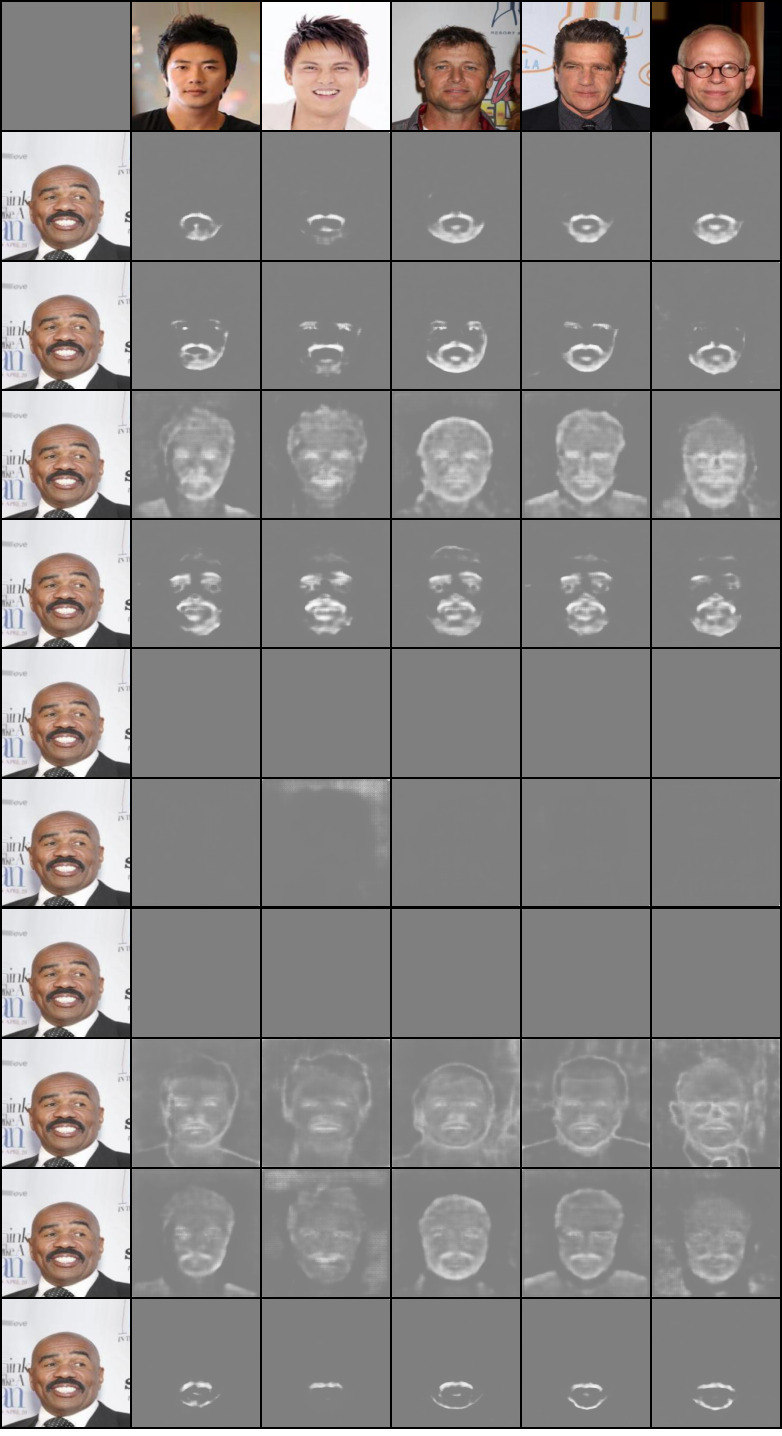}
    \caption{ Ablation analysis. The first row is images without facial hair on which we want to transfer the facial hair of the image in the first column. 
    Second row: All losses $\mathcal{L}$, third row: without $\mathcal{L}^A_{Recon2}$, fourth row: without $\mathcal{L}^B_{Recon2}$, fifth row: without $\mathcal{L}_{Cycle}$, sixth row: without $\mathcal{L}^B_{Recon1}$, seventh row: without $\mathcal{L}^A_{Recon1}$, eighth row: without $\mathcal{L}_{DC}$. The ninth row shows the translation where $\mathcal{L}^A_{Recon2}$ and $\mathcal{L}^B_{Recon2}$ are replaced by L2 regularization of the mask. The tenth row: L1 norm is replaced with L2 norm for $\mathcal{L}^A_{Recon2}$ and $\mathcal{L}^B_{Recon2}$. The last row: L1 norm is replaced with L2 norm for $\mathcal{L}^A_{Recon1}$ and $\mathcal{L}^B_{Recon1}$. }
    \label{fig:ablation}
\end{figure*}

Fig~\ref{fig:ablation} shows the masks generated when different losses are removed as discussed in the ablation analysis of section 4.3 of the main text.

\section{{  hyperparameters sensitivity}}
\label{sec:sensitivity}
\begin{figure*}[t]
    \centering
    \includegraphics[width=0.95\linewidth]{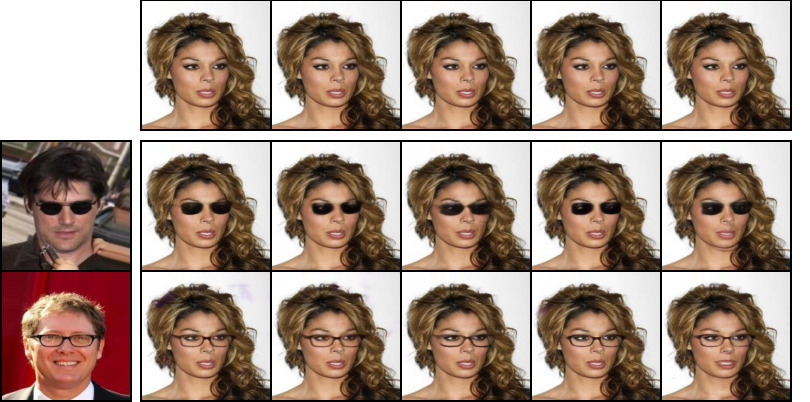}
    \caption{ Sensitivity to changes in $\lambda_4$. Given an image with glasses (left), and another image of a face with no glasses (top), the proposed method translates the specified glasses using different values of $\lambda_4$: $0.4,0.5,0.7,0.9,1.0$ (from left to right). All other hyperparameters remain fixed.} 
    \label{fig:lambda4}
\end{figure*}

\begin{figure*}[t]
    \centering
    \includegraphics[width=0.95\linewidth]{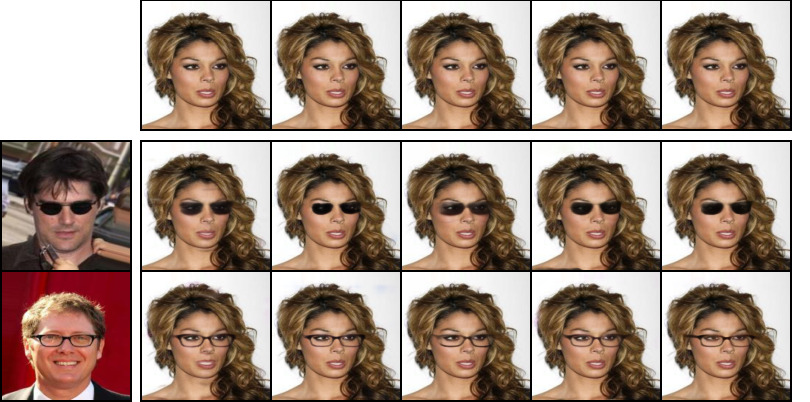}
    \caption{ Sensitivity to changes in $\lambda_1$. Given an image with glasses (left), and another image of a face with no glasses (top), the proposed method translates the specified glasses using different values of $\lambda_1$: $3.0,4.0,5.0,6.0,7.0$ (from left to right). All other hyperparameters remain fixed.} 
    \label{fig:lambda1}
\end{figure*}

The $\lambda$ coefficients were set in a way that reflects their relative importance and were observed. For example, if the mask obtained was too large we would increase $L_{Recon2}^A$ and $L_{Recon2}^B$. 
As illustrated in Fig.~\ref{fig:lambda4} and Fig.~\ref{fig:lambda1}, our network is not overly sensitive to the choice of these values. For example, for $L_{Recon2}^{A}$, each value in the range 0.4-1.0 results in a similar output and for $L_{Recon1}^{A}$ each value in the range 3.0-7.0 results in a similar output.

\begin{figure*}[t]
    \centering
    \includegraphics[width=0.95\linewidth]{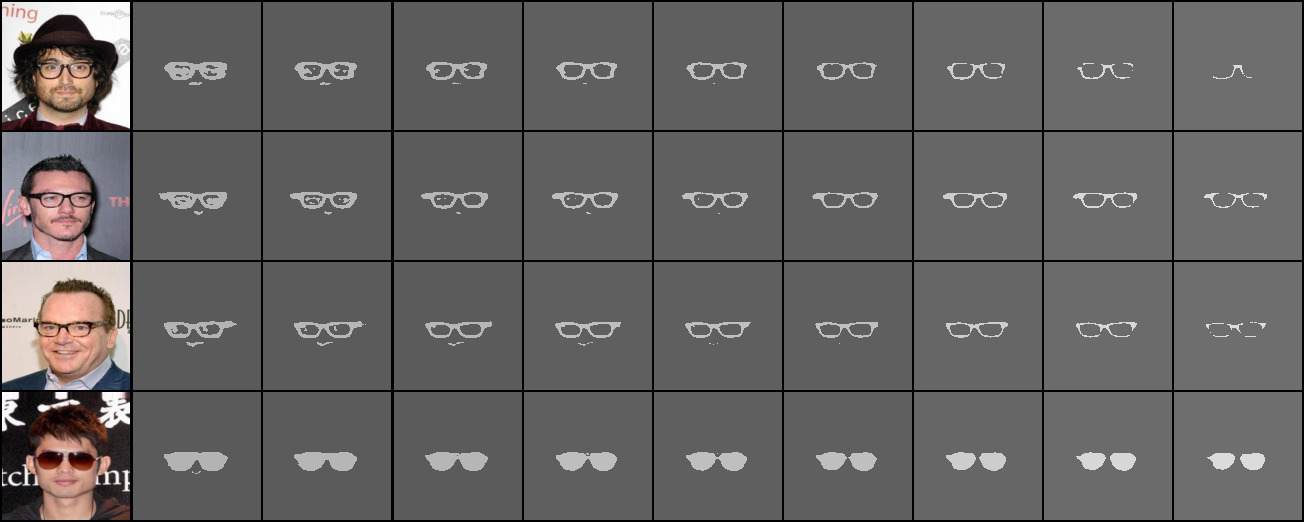}
    \caption{ The effect of the threshold used to binarized the mask. The left column is the original image and rest of the column are the segmentation mask created using the thresholds: $0.1,0.2,0.3,0.4,0.5,0.6,0.7,0.8,0.9$ (from left to right).} 
    \label{fig:threshold}
\end{figure*}

 Fig.~\ref{fig:threshold} shows the affect of the threshold used to binarized the mask for the task of segmentation. Fig.~\ref{fig:lambda4} and Fig.~\ref{fig:lambda1} show that the network is not overly sensitive to the choice of  $\lambda_4$ and $\lambda_1$ respectively.

\end{document}